\documentclass{article} 
\usepackage{iclr2027_conference,times}
\usepackage{times}

\usepackage{amsmath,amsfonts,bm}

\def\eqref#1{equation~\ref{#1}}

\def\1{\bm{1}}

\DeclareMathAlphabet{\mathsfit}{\encodingdefault}{\sfdefault}{m}{sl}
\SetMathAlphabet{\mathsfit}{bold}{\encodingdefault}{\sfdefault}{bx}{n}

\usepackage{url}
\usepackage{multirow}

\usepackage{wrapfig}
\definecolor{mydarkred}{rgb}{0.6,0,0}
\definecolor{mydarkgreen}{rgb}{0,0.6,0}
\usepackage[colorlinks,
linkcolor=mydarkred,
citecolor=blue]{hyperref}
\usepackage{soul} 
\usepackage{hyperref}
\usepackage{pifont}
\usepackage{url}
\usepackage[pdftex]{graphicx}
\usepackage{xspace}
\usepackage{booktabs}
\usepackage{amsmath}
\usepackage{amssymb}
\usepackage[table]{xcolor}  
\usepackage{subcaption} 
\definecolor{fix}{RGB}{0, 102, 204}
\usepackage{cancel}
\usepackage{bbm}

\usepackage{tcolorbox}
\tcbuselibrary{most}  
\usepackage{boxedminipage}
\usepackage{relsize}
\usepackage[dvipsnames]{xcolor}
\usepackage[dvipsnames]{xcolor}
\usepackage{minitoc}

\newcommand{\method}{WildTrace}

\title{\method: Evaluating Training Data Detection in Realistic Large-Scale LLM Post-Training}

\author{
Hengxiang Zhang\textsuperscript{1},\enspace
Junxian Liu \textsuperscript{1},\enspace
Hongxin Wei \textsuperscript{1}\thanks{Corresponding author} \\
\textsuperscript{1}Department of Statistics and Data Science, Southern University of Science and Technology \\
}

\iclrfinalcopy
\makeatletter
\newcommand{\promptfontsize}{\@setfontsize\promptfontsize{8pt}{8pt}} 
\makeatother

\newtcolorbox[auto counter, number within=section]{examplebox}[2][]{
  colback=white,
  colframe=blue!45!black,
  boxrule=0.8pt,
  rounded corners,
  left=4pt,
  right=4pt,
  top=4pt,
  bottom=4pt,
  enhanced,
  breakable, 
  title=Example~\thetcbcounter: #2,
  fonttitle={\bfseries\normalsize},  
  #1
}

\begin{document}

\doparttoc 
\faketableofcontents 

\maketitle

\begin{abstract}
Although post-training significantly enhances model performance, the use of massive undisclosed data raises concerns about benchmark contamination and copyright infringement.
This drives growing interest in training data detection for post-trained models.
Existing evaluations for training data detection often leverage benchmarks constructed post hoc, which can introduce distribution shifts. 
Alternatively, controlled evaluations rely on small-scale training setups that diverge from realistic post-training settings. 
As a result, their reported performance may not faithfully reflect the effectiveness of detection methods in real-world scenarios. 
To this end, we introduce \textsc{\method}, a reliable benchmark designed to evaluate detection methods in realistic, large-scale post-training settings.
Built on a multi-stage pipeline, \method\space encompasses stage-specific models throughout the post-training lifecycle, spanning SFT, DPO, and RL.
To perform a reliable evaluation, we construct model-specific evaluation sets across four diverse tasks, featuring distribution-matched members and non-members with verifiable membership labels.
Our results reveal that even the best-performing method achieves an average AUC of only 0.558 on our benchmark, which approaches the random-guessing level (0.5).
Furthermore, our analysis suggests that evaluation protocols can significantly affect measured detection performance. 
These findings highlight the significant potential to develop effective, practical, and robust detection methods in realistic scenarios. Our project resources are available at \url{https://jaydencoolcc.github.io/wildtrace-website/}.

\end{abstract}
\section{Introduction}
Modern Large Language Models (LLMs) exhibit remarkable capabilities across diverse tasks, owing to advances in post-training techniques~\citep{kumar2026llm, shao2024deepseekmath, yang2025qwen3}. 
The post-training stage typically involves large-scale Supervised Fine-Tuning (SFT), Direct Preference Optimization (DPO), and Reinforcement Learning (RL).
Although multi-stage post-training enhances model performance, the use of 
massive training data raises critical concerns about benchmark contamination~\citep{xu2024benchmark, choi2025contamination} and copyright infringement~\citep{slack2025early, li2026data}.
This underscores the need to audit training data from post-training stages.
However, the effectiveness of existing detection methods under realistic, large-scale post-training settings remains underexplored.

Previous work often relies on benchmarks constructed post hoc (e.g., WikiMIA) for evaluation~\citep{liu2024probing, shidetecting}. 
However, this post-hoc construction can introduce distribution shifts between members and non-members, compromising the validity of evaluation~\citep{duan2024membership, das2024blind}.
Although recent work mitigates this by employing distribution-matched members and non-members, their experimental setups diverge from realistic post-training settings.
Specifically, these controlled evaluations typically perform multi-epoch training on a small, domain-specific dataset, whereas realistic post-training leverages large-scale, diverse data for only one or a few epochs~\citep{fu2024membership,olmo2025olmo}.
Additionally, existing work often employs in-house evaluation with different models, datasets, and evaluation protocols, hindering fair comparison and reproducibility.
Therefore, a standardized benchmark is required to reliably evaluate training data detection under large-scale post-training scenarios.


To conduct a practical and reliable evaluation, an ideal benchmark for post-training data detection should adhere to three key principles: 
\textit{1) Distribution-matched sampling with verifiable membership:} Members and non-members are drawn from the same distributions with verifiable membership, avoiding distribution shifts and unreliable membership labels in post-hoc constructions.
\textit{2) Realistic large-scale post-training setup:} Target models are trained on diverse, large-scale datasets using standard post-training practices, establishing an evaluation testbed that faithfully reflects real-world deployment scenarios.
\textit{3) Reproducible evaluation across post-training lifecycle:} The benchmark provides publicly available evaluation suites consisting of target models and model-specific evaluation sets at each training stage, ensuring reproducible evaluation.

To this end, we introduce \method, a reliable benchmark for evaluating training data detection in realistic, large-scale post-training settings.
Leveraging the fully open Olmo-3 model, we perform a large-scale, multi-stage post-training pipeline (\textit{SFT} $\rightarrow$ \textit{DPO}$\rightarrow$ \textit{RL}).
This produces three stage-specific post-trained models, each with transparent training data and verifiable membership labels. 
To perform reliable evaluation, we construct model-specific evaluation sets across diverse tasks (Math, Coding, IF, and QA), where members and non-members are sampled from the same underlying distribution.
To facilitate research under realistic settings, our benchmark provides a reproducible testbed consisting of publicly available post-trained models and evaluation sets.

Using \method, we conduct a comprehensive evaluation of training data detection methods across the LLM post-training lifecycle. 
Our evaluation reveals that under large-scale post-training settings, existing detection methods exhibit limited effectiveness across all three stages—SFT, DPO, and RL.
Our analysis further indicates that scaling up training data generally degrades detection performance.
Consequently, high detection performance observed in small-scale training settings does not reflect actual performance under large-scale post-training settings.
This evidence highlights the need for developing more effective detection methods under large-scale post-training scenarios.

Beyond these findings, we revisit how evaluation protocols shape the assessment of training data detection.
First, empirical results indicate that evaluating training data detection on benchmarks with distribution shift may inflate detection performance, obscuring their actual performance in the wild.
Second, achieving high detection accuracy often requires calibrating decision thresholds with an in-distribution validation set, which is rarely available in real-world deployment.
Additionally, while reference-based methods can improve detection performance, they still fall short of effective detection in large-scale training settings.
Overall, these findings reveal a significant gap between current evaluation practices and realistic deployment, highlighting the need for more robust detection methods and practical evaluation protocols.







\section{Preliminaries}
\subsection{Training Data Detection for LLMs}
Training data detection for LLMs is an instance of membership inference~\citep{shokri2017membership, shidetecting}, which aims to identify whether a given text is used during model training.
In contrast to pretraining data detection, which identifies data in the pre-training corpus, post-training data detection aims to determine whether a sample is used at a particular post-training phase.
Let $\mathcal{M}$ denote the model post-trained on the dataset $\mathcal{D}$. 
Given a piece of text $x$, the objective is to predict whether the datapoint belongs to the training set. 
A detection method computes a membership score $S(x;\mathcal{M})$ derived from the predicted token probability of the model and makes a binary prediction:
\begin{equation}
h(x;\mathcal{M})=
\begin{cases}
1, & \text{if } S(x;\mathcal{M}) < \lambda,\\
0, & \text{if } S(x;\mathcal{M}) \geq \lambda,
\end{cases}
\end{equation}
where $h(x;\mathcal{M}) = 1$ indicates \textit{member} and $h(x;\mathcal{M}) = 0$ indicates \textit{non-member}, with $\lambda$ being a decision threshold.

\subsection{LLMs post-training}
Building on a pre-trained model, developers typically refine its behavior through a multi-stage post-training process.
Modern LLM post-training recipes typically involve three key paradigms: supervised fine-tuning (SFT), direct preference optimization (DPO), and reinforcement learning (RL). 
We introduce three widely used LLM post-training techniques in detail. 

\paragraph{Supervised Fine-Tuning}
Supervised Fine-Tuning aims to teach models to follow human instructions and generate responses with the desired format and style. 
To achieve this goal, a pre-trained model is further trained on a constructed high-quality dataset $\mathcal{D}$. 
Let $q$ denote a question and $y = \{y_1, y_2, \ldots, y_N\}$ denote the corresponding target response.
Using \(q\) as a prompt, SFT explicitly teaches the model to imitate the well-crafted target response. Formally, the training objective can be formulated as
\begin{equation}
    \mathcal{L}_{\theta}
    =
    -\,\mathbb{E}_{(q,y)\sim\mathcal{D}}
    \sum_{t=1}^{N}
    \log \pi_{\theta}
    \left(
        y_t \mid y_{<t}, q
    \right),
\end{equation}
where $\pi_\theta(y_t \mid y_{<t}, q)$ denotes the model's predicted probability for token $y_t$, given the preceding tokens. In practice, the SFT loss is typically computed only over response tokens, with the prompt serving as context.

\paragraph{Direct Preference Optimization}
Direct preference optimization is typically applied after SFT to further align the model with human preferences, using a preference dataset $\mathcal{D}$ consisting of triplets $(q, y^+, y^-)$. Concretely, DPO directly optimizes the policy to assign a higher likelihood to the preferred response $y^+$ over a less-preferred response $y^-$, which explicitly encodes the human's preference in models' parameters. Formally, the DPO loss is defined as follows:
\begin{equation}
\mathcal{L}^{\mathrm{DPO}}(\theta)
=
-\mathbb{E}_{(q,y^+,y^-)\sim\mathcal{D}}
\left[
\log \sigma
\left(
\beta \log
\frac{\pi_\theta(y^+ \mid q)}
     {\pi_{\mathrm{ref}}(y^+ \mid q)}
-
\beta \log
\frac{\pi_\theta(y^- \mid q)}
     {\pi_{\mathrm{ref}}(y^- \mid q)}
\right)
\right],
\end{equation}
where $\pi_\theta$ denotes the trainable policy, $\pi_{\mathrm{ref}}$ denotes a reference policy, and $\sigma(\cdot)$ is the sigmoid function. The hyperparameter $\beta$ controls the deviation of $\pi_\theta$ from $\pi_{\mathrm{ref}}$.

\paragraph{Reinforcement Learning} 
Group Relative Policy Optimization (GRPO)~\citep{shao2024deepseekmath} is a widely used RL algorithm for LLM training.
GRPO improves training efficiency over standard PPO by eliminating the critic model.
Specifically, for each question $q$, GRPO samples a group of $G$ responses $\{o_1, o_2, \ldots, o_G\}$ from the old policy $\pi_{\theta_{old}}$, and estimates each response's advantage from the relative rewards within the group. 
Given the reward $r_i$ for each response $o_i$, the advantage is computed as:
\begin{equation}
\hat{A}_{i,t} = \frac{r_i - \mathrm{mean}(\{r_1, r_2, \ldots, r_G\})}{\mathrm{std}(\{r_1, r_2, \ldots, r_G\})}.
\end{equation}
By using $\hat{A}_{i,t}$ as the advantage estimate, GRPO then minimizes a loss combining a clipped surrogate term with a KL penalty toward a reference policy $\pi_{ref}$:
\begin{align}
\mathcal{L}_\theta = -\,\mathbb{E}_{q,\, \{o_i\}_{i=1}^G} \frac{1}{G}\sum_{i=1}^{G}\frac{1}{|o_i|}\sum_{t=1}^{|o_i|}\Big[ &\min\big(\rho_{i,t}(\theta)\,\hat{A}_{i,t},\ \mathrm{clip}(\rho_{i,t}(\theta), 1-\epsilon, 1+\epsilon)\,\hat{A}_{i,t}\big) \notag \\
&- \beta\, \mathbb{D}_{\mathrm{KL}}\big[\pi_\theta \,\|\, \pi_{ref}\big] \Big],
\end{align}
where $\rho_{i,t}(\theta) = \frac{\pi_\theta(o_{i,t} \mid q, o_{i,<t})}{\pi_{\theta_{old}}(o_{i,t} \mid q, o_{i,<t})}$ is the importance sampling ratio, $\epsilon$ is the clipping threshold, and $\beta$ controls the strength of the KL regularization. 

\section{\method: Benchmarking Training Data Detection from Post-Trained LLMs}
Existing evaluations for training data detection often rely on post-hoc benchmarks or small-scale training settings, which fail to reflect the effectiveness of detection methods in the wild.
To bridge this gap, our work aims to systematically evaluate training data detection under more realistic settings.
In this section, we first identify the design principles of a reliable benchmark. 
We then introduce \method, a comprehensive benchmark for reliable evaluation of training data detection.

\subsection{Benchmark Design Principles}
\paragraph{Distribution-matched sampling with verifiable membership}
Conventional evaluation protocols often rely on post-hoc benchmark construction, treating documents published before a model's training cutoff date as members and those published after as non-members~\citep{shidetecting, yi2026membership, liu2024probing}.
However, temporal splits can introduce inadvertent artifacts (e.g., timestamps in WikiMIA) that cause distributional shifts between members and non-members~\citep{duan2024membership, das2024blind}.
Detection methods may exploit these artifacts rather than genuine membership signals, undermining the validity of the benchmark.
\textit{A reliable benchmark requires two requirements: (1) members and non-members are drawn from the same data distributions, and (2) membership labels are 
verifiable rather than heuristically inferred post hoc.}


\paragraph{Realistic large-scale post-training settings}
To mitigate distribution shifts, controlled evaluation protocols train target models on tailored datasets to obtain ground-truth membership labels~\citep{fu2024membership, tao2026detecting}.
However, these evaluation protocols commonly use small, domain-specific datasets trained for multiple epochs, whereas practical LLM training typically uses large-scale, diverse datasets with only one or a few epochs.
Small training corpora and repeated exposure can strengthen the model's memorization of individual samples, yielding optimistic detection performance.
Such idealized evaluations may overestimate the detectability of training data and do not reflect performance of detection methods under realistic post-training settings.
\textit{Consequently, a rigorous benchmark for training data detection should employ target models trained under realistic, large-scale training settings.}

\paragraph{Reproducible evaluation across post-training lifecycle}
While pre-training data detection has been extensively explored~\citep{shidetecting, zhang2024fine}, existing post-training studies typically focus on a single stage~\citep{fu2024membership, tao2026detecting}, such as supervised fine-tuning.
These works use diverse models, datasets, and evaluation protocols and rarely release their evaluation artifacts (e.g., trained models), hindering reproducibility and fair comparison.
Furthermore, post-training typically involves a multi-stage pipeline (\textit{SFT} $\rightarrow$ \textit{DPO} $\rightarrow$ \textit{RL}), necessitating comprehensive evaluation throughout post-training lifecycle.
Existing benchmarks fall short of providing a unified, reproducible framework for evaluating training data detection across the post-training lifecycle of LLMs.
\textit{A rigorous benchmark should provide publicly available oracle models and evaluation sets for each post-training stage, enabling reproducible evaluation.}

\subsection{Benchmark Curation}
Building on the proposed three principles, we construct \method\space by executing a realistic, large-scale post-training pipeline from scratch, which produces trained target models and corresponding evaluation sets at each stage.



\paragraph{Multi-stage oracle model training}
We build on the \textsc{Olmo-3} family~\citep{olmo2025olmo}, a fully open model family that releases models' training data, at a scale competitive with modern open LLMs.
This transparency allows us to construct a benchmark with verifiable membership labels under a large-scale post-training setting.
Starting from the base checkpoint \textit{Olmo-3-1025-7B}, we perform a large-scale, multi-stage post-training pipeline (\textit{SFT} $\rightarrow$ \textit{DPO} $\rightarrow$ \textit{RL}), with each stage trained from the model of the preceding stage.
We execute large-scale training using a diverse dataset of over 2.5 million examples, spanning 2.15M question-response pairs for SFT, 260K preference pairs for DPO, and 170K prompts for RL, conducted on a cluster of 32 NVIDIA A100 GPUs.
Through this construction process, we obtain three post-trained models, denoted \textsc{Oracle-SFT}, \textsc{Oracle-DPO}, and \textsc{Oracle-RL}, which serve as oracle models for evaluating detection methods.
More details of the model training are provided in Appendix~\ref{model_training}.

\paragraph{Model-specific evaluation set construction}
To construct a benchmark spanning the full post-training lifecycle, we establish a dedicated detection testbed using the models produced at each post-training stage.
Considering the diversity of training corpora, we curate four tasks for our benchmark: mathematics (Math), coding (Code), instruction following (IF), and general question answering (QA).
Specifically, for each task, we first select 250 samples from the original dataset as non-members, which are excluded from the training process.
The remaining examples constitute the training set, from which we further sample 250 samples per task as members. 
This guarantees that members and non-members are drawn from the same underlying distribution and that membership labels are verifiable.
Consequently, this process yields three evaluation sets paired with their oracle models: \textsc{\method-SFT} paired with \textsc{Oracle-SFT}, \textsc{\method-DPO} paired with \textsc{Oracle-DPO}, and 
\textsc{\method-RL} paired with \textsc{Oracle-RL}, each set containing four tasks.
The details of the benchmark are provided in Appendix~\ref{dataset}.

\subsection{Validating our \method\space with Blind Baselines}
\label{Validat_bench}
\begin{figure}[t]
    \centering
    \begin{subfigure}{0.32\linewidth}
        \centering
        \includegraphics[width=\linewidth]{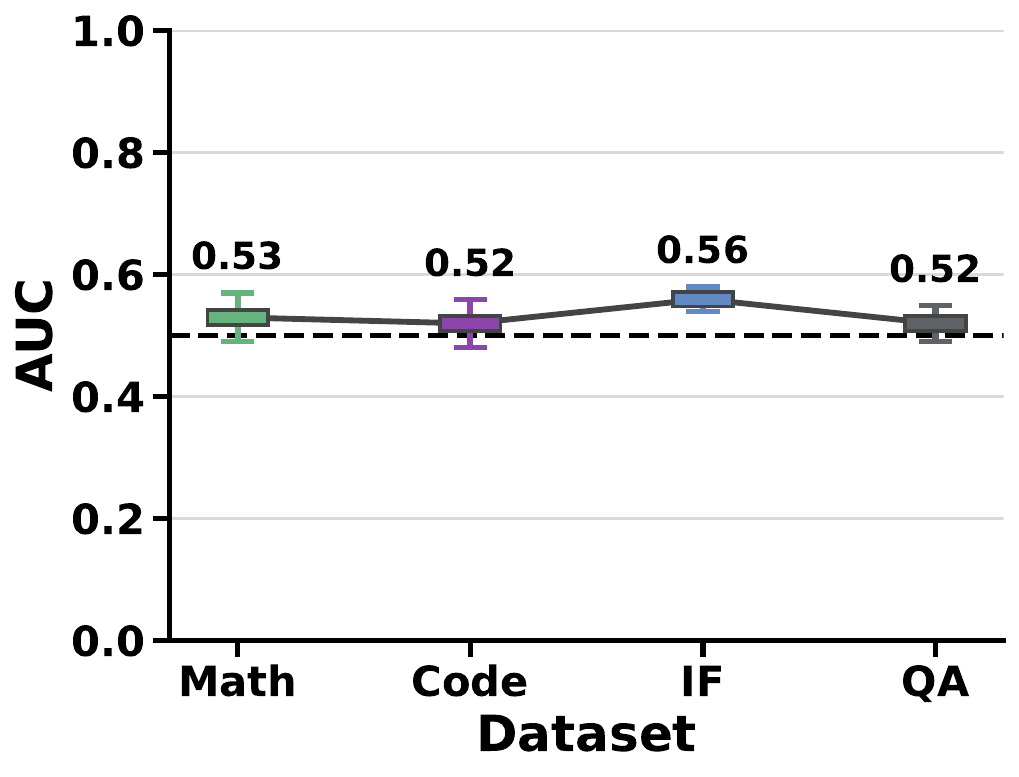}
        \caption{\textsc{\method-SFT}}
        \label{fig:best_auc_sft}
    \end{subfigure}
    \begin{subfigure}{0.32\linewidth}
        \centering
        \includegraphics[width=\linewidth]{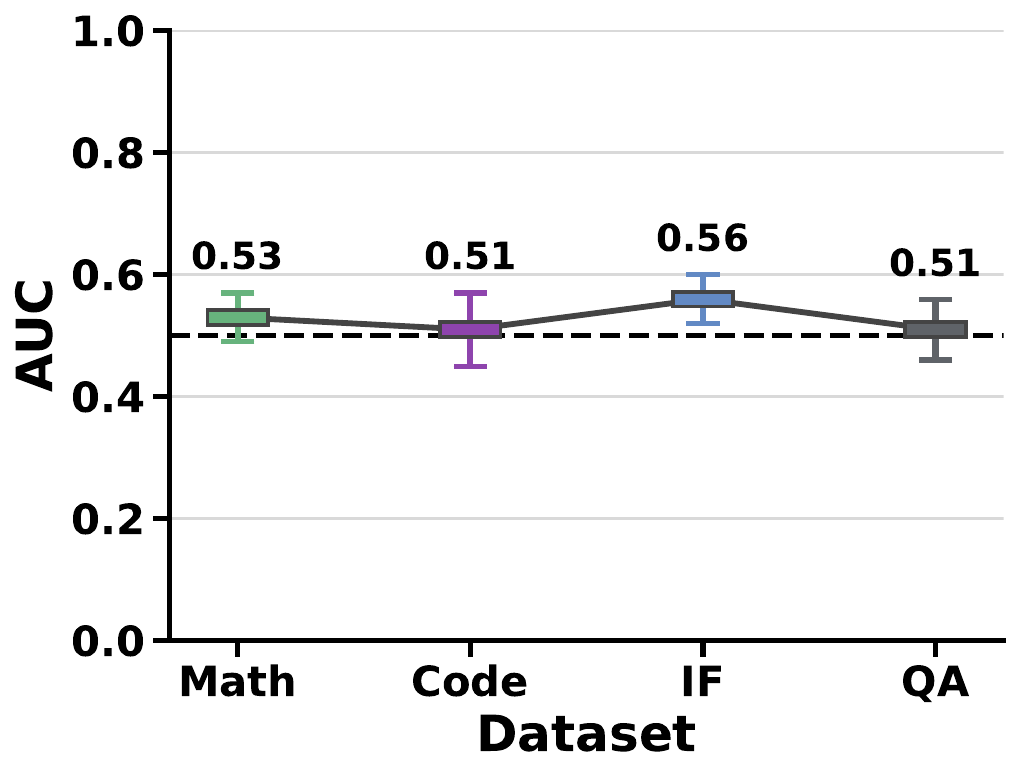}
        \caption{\textsc{\method-DPO}}
        \label{fig:best_auc_dpo}
    \end{subfigure}
    \begin{subfigure}{0.32\linewidth}
        \centering
        \includegraphics[width=\linewidth]{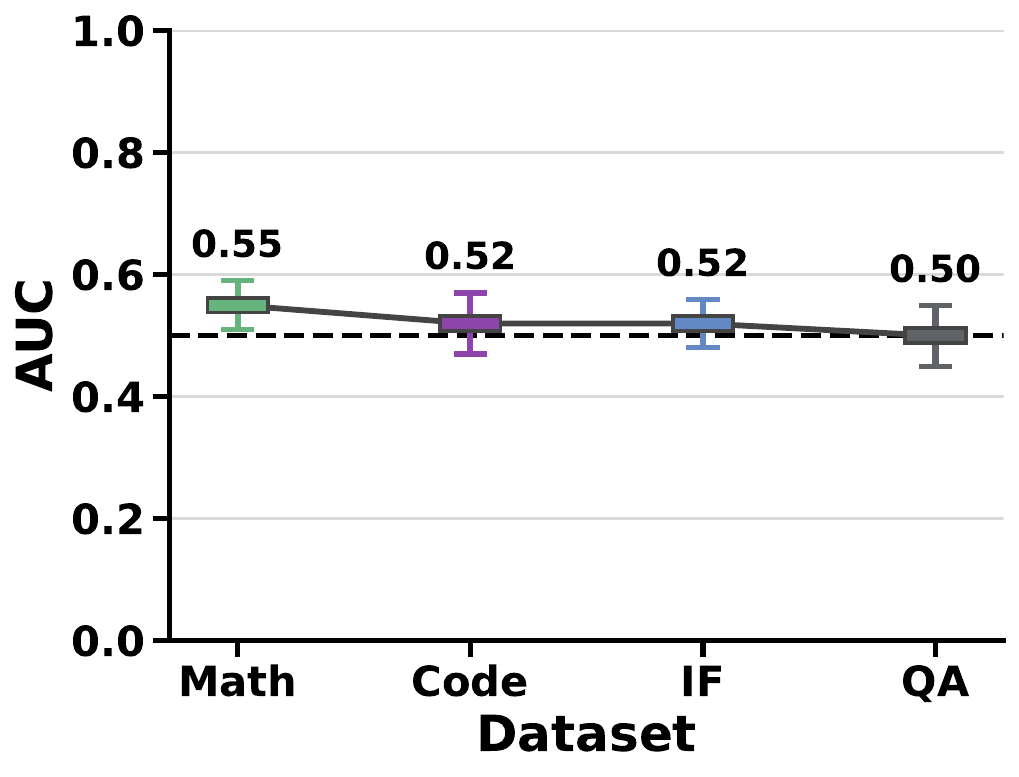}
        \caption{\textsc{\method-RL}}
        \label{fig:best_auc_rl}
    \end{subfigure}
\caption{The AUC of blind baselines across tasks and post-training stages.
For each task across post-training stages (a--c), we report the highest AUC achieved among nine blind baselines.
The dashed line indicates chance-level performance ($\text{AUC} = 0.50$). 
The near-random detection performance across all evaluation sets indicates the rigor and reliability of our benchmark construction.
}
\label{fig:best_auc_stages}
\end{figure}

To verify the reliability of \method, we test whether members and non-members can be distinguished using blind baselines that rely solely on text features and do not access the target model~\citep{das2024blind, wang2026checkmiabench}.
Specifically, we evaluate blind baseline approaches by combining three text feature extractors (Bag-of-Words, TF-IDF, Word2Vec) with three classifiers (Logistic Regression, Random Forest, Neural Network), resulting in nine attack combinations. 
Figure~\ref{fig:best_auc_stages} shows the best AUC among the nine blind baselines across different post-training stages.
The blind baselines exhibit near-random performance, with average AUCs across tasks of 0.53, 0.53, and 0.52 on \textsc{\method-SFT}, \textsc{\method-DPO}, and \textsc{\method-RL}, respectively.
These results show that blind baselines obtain negligible signals for distinguishing members from non-members on our benchmark, validating the reliability of our benchmark. 
We provide additional experimental results in Appendix~\ref{no_shift}, further confirming the validity of our benchmark.

\section{Experiment}
\subsection{Experimental setup}
\paragraph{Detection methods}
To investigate the performance of existing detection methods on our benchmark, we evaluate commonly used detection methods for LLMs.
We categorize these methods into two classes based on their scoring signal: \textit{input-based methods} and \textit{output-based methods}. Specifically, \textit{input-based methods} compute membership scores based on the likelihood of input tokens, including \textit{Perplexity}~\citep{li2023estimating}, \textit{Entropy}~\citep{song2021systematic}, 
\textit{Zlib}~\citep{carlini2021extracting}, \textit{Lowercase}~\citep{carlini2021extracting},
\textit{Neighbor}~\citep{mattern2023membership}, \textit{Min-K\%}~\citep{shidetecting}, and \textit{Min-K\%++}~\citep{zhang2024min}.
\textit{Output-based methods} leverage tokens generated by the model as the detection signal, including \textit{CDD}~\citep{dong-etal-2024-generalization}, \textit{Min-kNN}~\citep{zhang2026detecting}, \textit{Self-Critique}~\citep{tao2026detecting}, \textit{Generated Perplexity}, \textit{Generated Entropy} and \textit{Generated Min-K\%}.
We provide detailed descriptions and implementation details for each method in Appendix~\ref{methods}.

\paragraph{Implementation details}
In our experiments, for \textit{input-based methods}, we compute membership scores on input tokens. For \textit{output-based methods}, we condition on the question to generate responses and compute scores on the generated tokens. 
Unless otherwise specified, we set the maximum sequence length to 2048 tokens. 
During the detection phase, we assume access to the complete information of each sample. Further evaluation details are provided in Appendix~\ref{expe:evaluation_details}.
We evaluate the effectiveness of existing detection methods on our \method\space by measuring the following metrics: (1) AUC, the area under the receiver operating characteristic curve; (2) TPR@$5\%\mathrm{FPR}$, the true positive rate at a $5\%$ false positive rate \citep{carlini2022membership}. Instead of paying equal attention to members and non-members, this metric pays more attention to members and evaluates whether one can confidently identify members.

\subsection{Main Results}
\label{main_result}
\paragraph{Detecting training data remains challenging under large-scale post-training scenarios.}
Table~\ref{tab:main_auc} reports the AUC of detection methods across diverse post-trained models from different post-training stages, including SFT, DPO, and RL training paradigms.
The experimental results show that most detection methods achieve AUC scores near random guess on our benchmark. 
The TPR@5\%FPR of the detection methods is provided in Table~\ref{tab:main_tpr} in Appendix~\ref{add_results}.
This finding indicates that existing detection methods struggle to effectively identify training data under large-scale post-training scenarios.
We present the results of reference-based methods in Appendix~\ref{sec:reference-based}, which yield consistent findings.
Compared to DPO and RL, we find that training data from the SFT stage is more detectable.
For example, the best-performing method \textsc{Min-K\%++} achieves an AUC of 0.758 on QA under \textsc{Oracle-SFT}, but only 0.544 and 0.491 under \textsc{Oracle-DPO} and \textsc{Oracle-RL}, respectively.
This gap may arise from differences in the training objectives, which rely on distinct supervision signals.
SFT maximizes the likelihood of target responses conditioned on the question, leaving a strong signal for detection. 
In contrast, DPO optimizes relative likelihood between preferred and less-preferred responses, while RL maximizes reward over its online sampled rollouts. 
Both objectives leave a weaker detectable signal, making their training data less amenable to detection.
Overall, these findings indicate that current detection methods exhibit limited effectiveness under large-scale post-training scenarios, necessitating more capable detection methods. 
\begin{table*}[!t]
\centering
\footnotesize
\caption{AUC of detection methods on our benchmark. Reported results are averaged over three runs with different random seeds. \textit{Input-based methods} compute scores using input tokens, whereas \textit{Output-based methods} score output tokens. Avg. denotes the AUC of each method, averaged across all tasks and post-trained models. \textbf{Bold} indicates the best result.}
\label{tab:main_auc}
\renewcommand{\arraystretch}{1.2}

\resizebox{\textwidth}{!}{%
\setlength{\tabcolsep}{1.5mm}%
\begin{tabular}{@{}lccccccccccccc@{}}
\toprule

\multirow{2}{*}{\textbf{Method}}
& \multicolumn{4}{c}{\textbf{\textsc{Oracle-SFT}}}
& \multicolumn{4}{c}{\textbf{\textsc{Oracle-DPO}}}
& \multicolumn{4}{c}{\textbf{\textsc{Oracle-RL}}}
& \multirow{2}{*}{\textbf{Avg.}} \\

\cmidrule(lr){2-5}
\cmidrule(lr){6-9}
\cmidrule(lr){10-13}

& Math & Code & IF & QA
& Math & Code & IF & QA
& Math & Code & IF & QA
& \\

\midrule

\rowcolor{gray!8}
\multicolumn{14}{l}{
    \textcolor{gray}{\textbf{\textit{Input-based methods}}}
} \\

\addlinespace[2pt]

Perplexity
& 0.537 & 0.570 & 0.620 & 0.660
& 0.500 & 0.451 & 0.527 & 0.514
& 0.530 & 0.492 & 0.503 & 0.475
& 0.532 \\

Entropy
& 0.501 & 0.520 & 0.486 & 0.444
& 0.524 & 0.447 & 0.487 & 0.488
& 0.515 & 0.492 & 0.510 & 0.474
& 0.491 \\

Lowercase
& 0.595 & 0.593 & 0.628 & 0.673
& 0.498 & 0.438 & 0.514 & 0.523
& 0.504 & 0.492 & \textbf{0.522} & 0.467
& 0.537 \\

Zlib
& 0.518 & 0.521 & 0.579 & 0.567
& 0.466 & 0.538 & 0.526 & 0.465
& 0.513 & \textbf{0.518} & 0.513 & \textbf{0.505}
& 0.519 \\

Neighbor
& 0.548 & 0.541 & 0.506 & 0.567
& 0.515 & 0.429 & 0.506 & \textbf{0.544}
& 0.516 & 0.502 & 0.486 & 0.487
& 0.512 \\

\textsc{MIN-K\%}
& 0.540 & 0.569 & 0.606 & 0.662
& 0.496 & 0.459 & 0.535 & 0.519
& 0.526 & 0.491 & 0.502 & 0.476
& 0.532 \\

\textsc{MIN-K\%++}
& \textbf{0.602} & \textbf{0.615} & \textbf{0.646} & \textbf{0.758}
& 0.504 & \textbf{0.543} & 0.473 & \textbf{0.544}
& 0.545 & 0.462 & 0.508 & 0.491
& \textbf{0.558} \\

\rowcolor{gray!8}
\multicolumn{14}{l}{
    \textcolor{gray}{\textbf{\textit{Output-based methods}}}
} \\

\addlinespace[2pt]

CDD
& 0.501 & 0.500 & 0.478 & 0.497
& 0.505 & 0.516 & 0.477 & 0.515
& 0.485 & 0.477 & 0.483 & 0.477
& 0.493 \\

Generated Perplexity
& 0.507 & 0.513 & 0.483 & 0.517
& 0.520 & 0.481 & 0.544 & 0.488
& \textbf{0.546} & 0.443 & 0.464 & 0.469
& 0.498 \\

Generated Entropy
& 0.503 & 0.513 & 0.479 & 0.520
& 0.521 & 0.481 & 0.534 & 0.488
& 0.543 & 0.444 & 0.465 & 0.470
& 0.497 \\

Generated \textsc{MIN-K\%}
& 0.509 & 0.516 & 0.484 & 0.515
& 0.519 & 0.480 & \textbf{0.547} & 0.498
& \textbf{0.546} & 0.444 & 0.462 & 0.476
& 0.500 \\

Min-kNN
& 0.517 & 0.532 & 0.479 & 0.520
& \textbf{0.525} & 0.452 & 0.512 & 0.528
& 0.530 & 0.430 & 0.458 & 0.471
& 0.496 \\

Self-Critique
& 0.566 & 0.543 & 0.487 & 0.473
& 0.484 & 0.473 & 0.509 & 0.509
& 0.483 & 0.493 & 0.514 & 0.462
& 0.500 \\

\bottomrule
\end{tabular}%
}
\end{table*}

\begin{figure}[t]
    \centering
    \begin{subfigure}{0.48\linewidth}
        \centering
        \includegraphics[width=\linewidth]{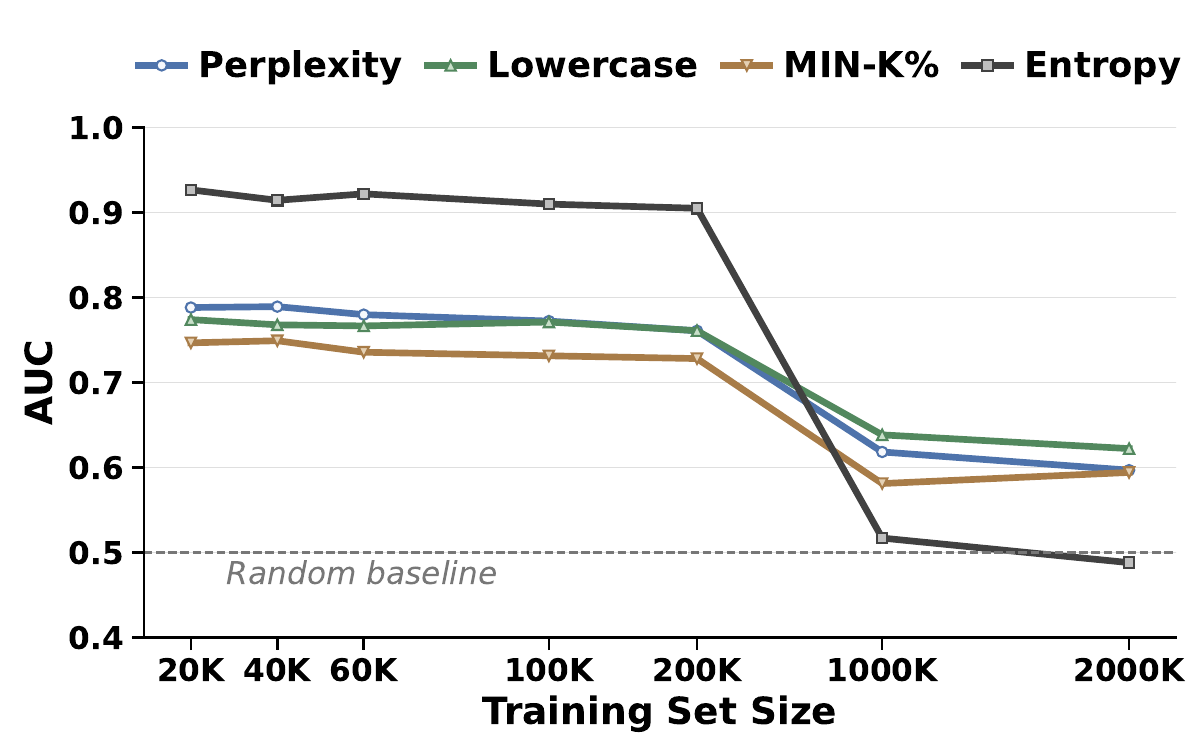}
        \caption{AUC vs. Training set size}
        \label{fig:auc_datasize}
    \end{subfigure}
    \hfill
    \begin{subfigure}{0.48\linewidth}
        \centering
        \includegraphics[width=\linewidth]{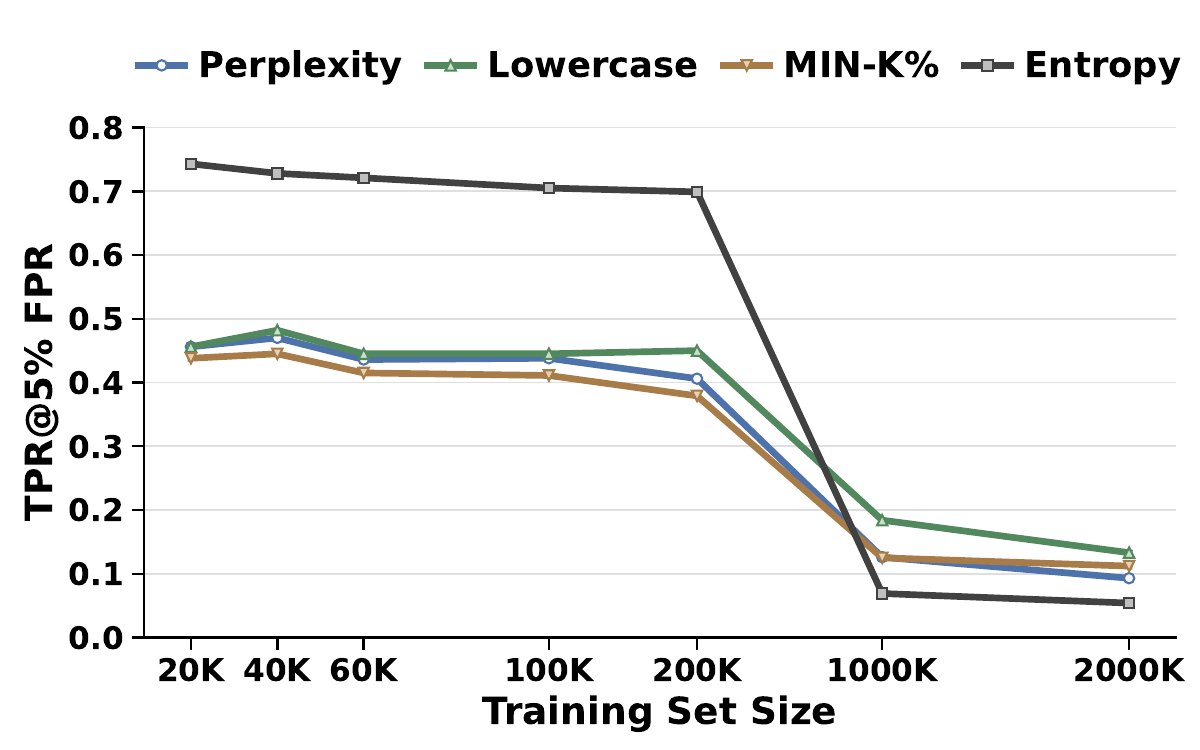}
        \caption{TPR@5\%FPR vs. Training set size}
        \label{fig:tpr_datasize}
    \end{subfigure}
\caption{Effect of training set size on detection performance. Both AUC and TPR@5\% FPR degrade as the training data scale increases, with AUC ultimately approaching the random baseline.}
\end{figure}

\paragraph{Strong detection performance in small-scale settings does not transfer to large-scale training.}
Existing training data detection evaluations are typically conducted in small-scale training settings, misaligned with realistic LLM post-training settings.
To examine how training scales affect detection performance, we train the base model on datasets ranging from 20K to 2000K examples, sampled from Dolci-Instruct-SFT~\citep{olmo2025olmo}. 
In our experiments, all training configurations are held constant except for the training set size. 
Across all training scales, we include all benchmark members in the training set and evaluate the performance of detection methods on \method.
Figures~\ref{fig:auc_datasize} and \ref{fig:tpr_datasize} report the average AUC and TPR@5\%FPR of various detection methods across four tasks, respectively.
The results show that detection performance degrades as training scales up.
For instance, the AUC of Entropy decreases from 0.927 to 0.517 as the training set size increases from 20K to 1000K, approaching the random-guessing level.
These findings reveal that the detection performance evaluated in small-scale training settings does not transfer to large-scale practical training settings.
This highlights the importance of evaluating detection methods in realistic, large-scale training settings, as our \method\space benchmark provides.

\section{Revisiting Training Data Detection from LLMs}
Training data detection has attracted growing attention as a measure for auditing data contamination, privacy, and model memorization in LLMs. 
However, the evaluation protocol substantially affects detection performance in practice.
In this section, we examine key evaluation settings that are critical for effectively detecting training data in realistic post-training scenarios.

\subsection{Impact of evaluation settings on detection performance}
\label{sec:Revisiting}
\paragraph{Setup}
Our experiments in Section~\ref{main_result} show that current detection methods exhibit limited detection performance under large-scale post-training settings. 
To better observe performance differences across diverse evaluation settings, we conduct experiments under small-scale training settings.
We fine-tune the Olmo 3 base model on 10K examples from Dolci-Instruct-SFT, resulting in {\textsc{Oracle-SFT-10K}}.
Detailed implementations are provided in Appendix~\ref{expe_revisiting}.
In this section, unless otherwise specified, our evaluations are conducted under this evaluation configurations.

\begin{figure}[t]
    \centering
    \begin{subfigure}{0.42\linewidth}
        \centering
        \includegraphics[width=\linewidth]{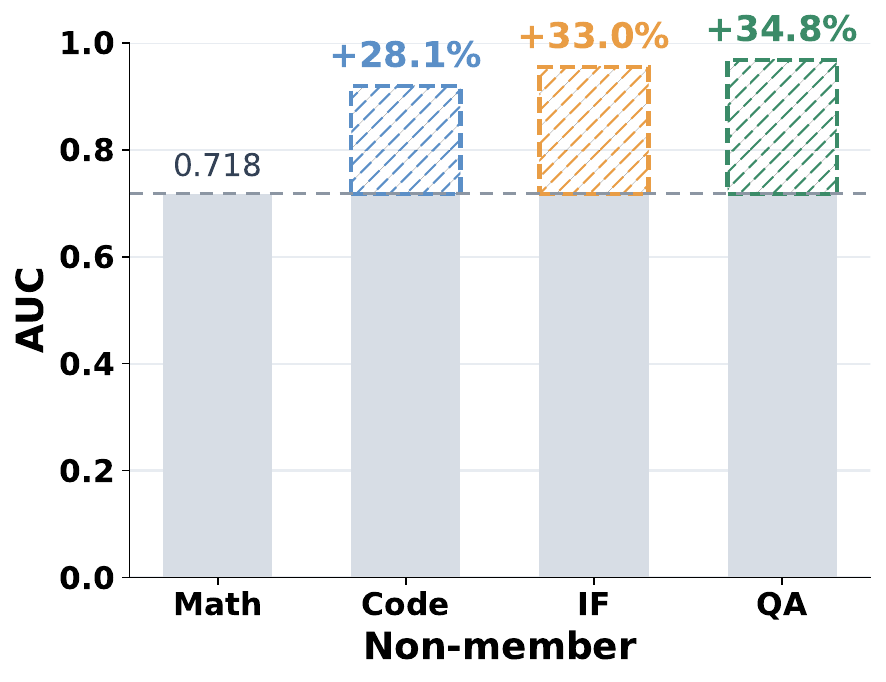}
        \caption{AUC across different non-member sets.}
        \label{fig:auc_cross_domain}
    \end{subfigure}
    \hspace{0.3mm}
    \begin{subfigure}{0.42\linewidth}
        \centering
        \includegraphics[width=\linewidth]{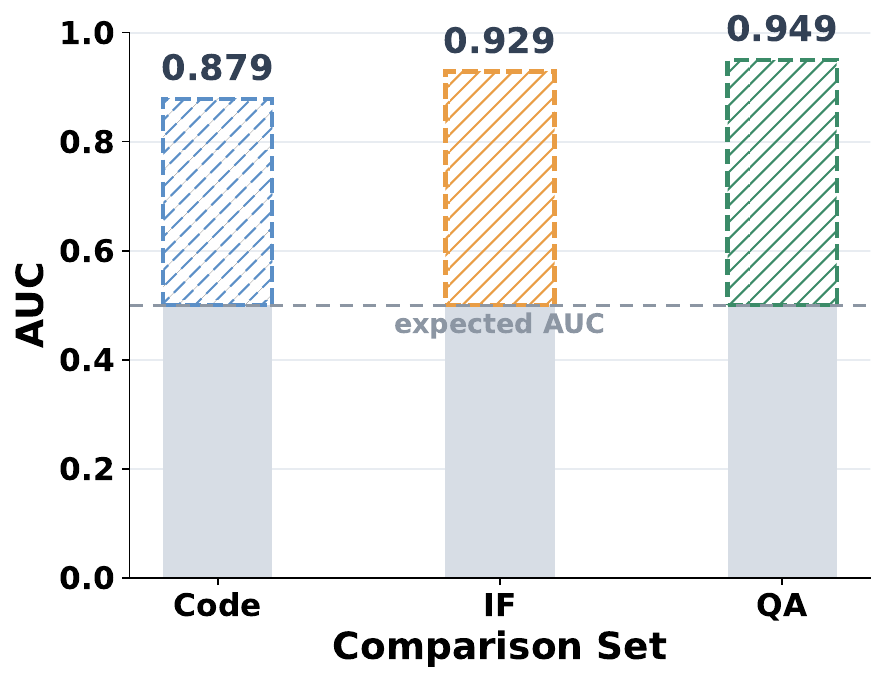}
        \caption{AUC between non-member sets.}
        \label{fig:auc_nonmembers}
    \end{subfigure}
\caption{The AUC scores of the perplexity-based method under distribution shift. 
(a) AUC for math members and non-members across domains. Percentages show inflated improvement relative to the in-domain math evaluation set.
(b) AUC across domains for non-member vs. non-member comparisons, using math as the reference domain.}
\end{figure}

\paragraph{The composition of evaluation benchmarks significantly affects evaluation results.}
Training data detection is commonly evaluated on a dataset consisting of members and non-members, yet the composition of this evaluation set can affect the evaluation results.
To examine this effect, we evaluate detection methods across evaluation sets with different distribution shifts between members and non-members.
Figure~\ref{fig:auc_cross_domain} shows the AUC of the detection method for detecting math training data when non-members come from different domains.
The results show that the detection method achieves inflated AUC scores on the math task when non-members are drawn from other domains, with performance gains of up to +34.8\% compared to in-domain evaluation. 
To further illustrate this effect, we construct evaluation sets where \textit{both} groups are non-members (i.e., neither was used for training).
As shown in Figure~\ref{fig:auc_nonmembers}, the detection method still achieves high AUC scores, reaching 0.949 for QA vs. Math, substantially above the chance level of 0.5.
This inauthentic detection performance stems from domain-specific distribution shifts between the two non-member groups, resulting in separable detection score distributions, as shown in Appendix~\ref{distribution_shift}.
These results indicate that distribution shifts enable detection methods to achieve superior performance by exploiting superficial artifacts rather than true membership signals.
However, post-hoc benchmark construction often cannot ensure that members and non-members are drawn from the same underlying distribution, which is not an ideal evaluation setup.
A reliable benchmark requires minimizing distribution shifts between members and non-members, as shown in Appendix~\ref{no_shift}.

\paragraph{Superior detection accuracy requires calibrating the decision threshold on an ID validation set.}
In a real-world scenario, deploying a detection algorithm often requires a preset decision threshold. 
Existing studies typically assume access to a validation set drawn from the same distribution as the evaluation set to calibrate this threshold.
To assess the impact of the validation set on detection accuracy, we compare three threshold calibration settings: an in-distribution (ID) validation set, an out-of-distribution (OOD) validation set, and no validation set (e.g., using BlindMI~\citep{BlindMI}).
Further details of the implementation are provided in Appendix~\ref{expe_threshold}.
The results in Table~\ref{tab:threshold} show that calibrating thresholds using an ID validation set can yield the best detection performance.
In contrast, calibrating thresholds using an OOD validation set degrades detection performance.
In the absence of a validation set, unsupervised threshold calibration methods achieve an average accuracy of up to 0.664, but remain inferior to supervised or oracle strategies.
These findings show that superior detection accuracy often requires calibrating the threshold with an ID validation set.
However, in practice, obtaining a validation set aligned with the distribution of the test set is rarely feasible.
This gap between the evaluation assumptions and the practical implementation motivates future research to focus on more effective threshold calibration strategies.
\begin{table}[!t]
\centering
\caption{The accuracy of detection methods on the Code task under different threshold calibration approaches. 
Thresholds are calibrated using ID validation data (Code) and OOD validation data (IF, QA, and Math), respectively. 
Each cell reports the mean $\pm$ standard deviation over 10 runs, where different random seeds yield different validation sets. 
\textbf{Bold} indicates the best result.}
\label{tab:threshold}
\renewcommand{\arraystretch}{1.3}
\setlength{\tabcolsep}{1.5mm}
\resizebox{\linewidth}{!}{
\begin{tabular}{llccccccc}
\toprule
\multicolumn{2}{c}{\textbf{Threshold Calibration}}
& \multicolumn{6}{c}{\textbf{Methods}}
& \multirow{2}{*}{\textbf{Avg.}} \\
\cmidrule(lr){1-2}\cmidrule(lr){3-8}

\textbf{Setting}
& \textbf{Set / Method}
& \textbf{Perplexity}
& \textbf{Entropy}
& \textbf{Lowercase}
& \textbf{Zlib}
& \textbf{\textsc{MIN-K\%}}
& \textbf{\textsc{MIN-K\%++}} \\
\midrule

\multirow{3}{*}{\textbf{OOD}}
& IF
& $0.674 \pm 0.03$
& $\mathbf{0.787 \pm 0.01}$
& $\underline{0.663 \pm 0.03}$
& $\underline{0.594 \pm 0.02}$
& $0.665 \pm 0.02$
& $0.739 \pm 0.04$
& $\underline{0.687}$ \\

& QA
& $\mathbf{0.709 \pm 0.02}$
& $0.778 \pm 0.02$
& $0.643 \pm 0.03$
& $0.578 \pm 0.04$
& $0.659 \pm 0.03$
& $0.666 \pm 0.09$
& $0.672$ \\

& Math
& $0.515 \pm 0.01$
& $0.666 \pm 0.02$
& $0.642 \pm 0.02$
& $0.506 \pm 0.00$
& $0.515 \pm 0.01$
& $0.679 \pm 0.05$
& $0.587$ \\

\midrule
\multirow{3}{*}{\textbf{Unsupervised}}
& GMM
& $0.606 \pm 0.00$
& $0.772 \pm 0.00$
& $0.622 \pm 0.00$
& $0.544 \pm 0.00$
& $\underline{0.682 \pm 0.00}$
& $0.728 \pm 0.00$
& $0.659$ \\

& GMM$_{\mathrm{log1p}}$
& $0.670 \pm 0.00$
& $0.766 \pm 0.00$
& $0.624 \pm 0.00$
& $0.544 \pm 0.00$
& $0.652 \pm 0.00$
& $0.730 \pm 0.00$
& $0.664$ \\

& BlindMI
& $0.684 \pm 0.00$
& $0.510 \pm 0.00$
& $0.642 \pm 0.00$
& $0.502 \pm 0.00$
& $0.502 \pm 0.00$
& $\mathbf{0.784 \pm 0.00}$
& $0.604$ \\

\midrule
\rowcolor{gray!15}
\textbf{ID}
& Code
& $\underline{0.704 \pm 0.02}$
& $\underline{0.785 \pm 0.02}$
& $\mathbf{0.668 \pm 0.02}$
& $\mathbf{0.603 \pm 0.02}$
& $\mathbf{0.688 \pm 0.01}$
& $\underline{0.764 \pm 0.02}$
& $\mathbf{0.702}$ \\
\bottomrule
\end{tabular}
}
\end{table}

\paragraph{Does the performance of the reference-based method rely on the choice of reference model?}
Reference-based detection methods typically achieve the strongest detection performance among existing approaches. 
However, they often assume access to the exact base model used to train the target model.
In practice, this assumption is rarely satisfied, as model developers typically do not disclose the base model of the post-trained model.
To investigate how the choice of reference model affects detection performance, we evaluate reference-based methods using a range of reference models (e.g., Qwen and Llama families).
Table~\ref{tab:auc_reference_model_ppl} compares the AUC of the reference-based method using different reference models and a reference-free baseline on \textsc{Oracle-SFT} models trained with 100K, 200K, and 1000K examples.
\textit{First, while reference-based methods can enhance detection performance, their effectiveness depends on the choice of the reference model.}
Specifically, on Oracle-SFT-1000K, the average AUC ranges from 0.68 to 0.74 across reference models, outperforming the reference-free baseline of 0.62.
\textit{Second, the reference-based method's performance degrades significantly as the training data scale increases}, with the average AUC decreasing from 0.89 to 0.86 and 0.70 on \textsc{Oracle-SFT-100K}, \textsc{Oracle-SFT-200K}, and \textsc{Oracle-SFT-1000K}, respectively, when using the exact base model (OLMo-3-1025-7B) as the reference model.
The additional results in Appendix~\ref{sec:reference-based} show that reference-based methods still exhibit limited effectiveness in detecting training data under large-scale post-training scenarios.
These results demonstrate that while reference-based methods benefit from a suitable reference model, they remain susceptible to performance degradation as the training data scale increases.

\begin{table*}[t]
\centering
\caption{AUC of the reference-based method using perplexity across
different reference models under 100K, 200K, and 1000K training data settings.
\textbf{Bold} indicates the superior results.}
\label{tab:auc_reference_model_ppl}

\renewcommand{\arraystretch}{1.1}
\setlength{\tabcolsep}{1.5mm}

\resizebox{\linewidth}{!}{%
\begin{tabular}{llccccc|ccccc|ccccc}
\toprule
\multicolumn{2}{c}{\textbf{Reference Models}}
& \multicolumn{5}{c|}{\textbf{\textsc{Oracle-SFT-100K}}}
& \multicolumn{5}{c|}{\textbf{\textsc{Oracle-SFT-200K}}}
& \multicolumn{5}{c}{\textbf{\textsc{Oracle-SFT-1000K}}} \\
\cmidrule(lr){1-2}
\cmidrule(lr){3-7}
\cmidrule(lr){8-12}
\cmidrule(lr){13-17}

\textbf{Model} & \textbf{Type}
& \textbf{IF} & \textbf{QA} & \textbf{Math} & \textbf{Coding} & \textbf{Avg.}
& \textbf{IF} & \textbf{QA} & \textbf{Math} & \textbf{Coding} & \textbf{Avg.}
& \textbf{IF} & \textbf{QA} & \textbf{Math} & \textbf{Coding} & \textbf{Avg.} \\
\midrule

Reference-free & \textendash
& 0.77 & 0.87 & 0.69 & 0.77 & 0.77
& 0.76 & 0.86 & 0.67 & 0.75 & 0.76
& 0.63 & 0.70 & 0.54 & 0.60 & 0.62 \\
\cmidrule(lr){1-17}

Olmo-3-7B-Instruct & Instruct
& 0.84 & 0.94 & 0.87 & 0.86 & 0.88
& 0.82 & 0.92 & 0.81 & 0.83 & 0.85
& 0.67 & 0.76 & 0.61 & 0.69 & 0.68 \\

Olmo-3-1025-7B & Base
& 0.86 & 0.94 & 0.89 & 0.87 & 0.89
& 0.84 & 0.92 & 0.83 & 0.84 & 0.86
& 0.72 & 0.78 & 0.61 & 0.71 & 0.70 \\
\cmidrule(lr){1-17}

Qwen2.5-7B-Instruct & Instruct
& 0.88 & 0.95 & 0.92 & 0.87 & 0.91
& 0.85 & 0.93 & 0.86 & 0.84 & 0.87
& 0.72 & 0.79 & 0.64 & 0.70 & 0.71 \\

Qwen2.5-7B & Base
& 0.86 & 0.94 & 0.90 & 0.86 & 0.89
& 0.84 & 0.92 & 0.84 & 0.84 & 0.86
& 0.72 & 0.78 & 0.62 & 0.69 & 0.70 \\
\cmidrule(lr){1-17}

Qwen3-1.7B & Instruct
& \textbf{0.89} & \textbf{0.96} & 0.93 & 0.88 & 0.91
& \textbf{0.86} & \textbf{0.94} & 0.87 & \textbf{0.85} & 0.88
& 0.74 & 0.80 & 0.63 & 0.73 & 0.73 \\

Qwen3-1.7B-Base & Base
& 0.87 & 0.95 & 0.91 & 0.87 & 0.90
& 0.84 & 0.93 & 0.84 & 0.84 & 0.86
& 0.73 & 0.79 & 0.63 & 0.70 & 0.72 \\
\cmidrule(lr){1-17}

Llama-3.2-3B-Instruct & Instruct
& 0.89 & 0.95 & 0.95 & \textbf{0.88} & \textbf{0.92}
& 0.86 & 0.93 & 0.88 & 0.85 & 0.88
& \textbf{0.75} & \textbf{0.80} & 0.67 & \textbf{0.74} & \textbf{0.74} \\

Llama-3.2-3B & Base
& 0.88 & 0.95 & \textbf{0.96} & 0.87 & 0.91
& 0.86 & 0.93 & \textbf{0.90} & 0.84 & \textbf{0.88}
& 0.75 & 0.78 & \textbf{0.67} & 0.72 & 0.73 \\

\bottomrule
\end{tabular}%
}
\end{table*}



\section{Conclusion}
In this work, we introduce \method, a reliable benchmark designed to evaluate training data detection under realistic, large-scale post-training settings. 
Our benchmark is built from scratch by performing large-scale post-training in the SFT, DPO, and RL stages, with evaluation sets drawn from the same data distributions to mitigate potential distribution shifts.
Our comprehensive evaluation demonstrates that current detection methods exhibit limited effectiveness at scale.
Importantly, we find that performance observed in small-scale settings does not generalize to large-scale post-training, underscoring the critical need to evaluate methods under realistic scenarios.
These findings motivate the development of detection approaches that are more effective, practical, and robust under modern post-training settings.
Overall, we hope \method\space can serve as a solid testbed for advancing reliable and deployable training data detection in LLM post-training.

\paragraph{Limitations}
Our benchmark aims to provide a practical and reliable testbed for research on training data detection.
To ensure accurate ground-truth membership without dataset contamination, our \method\space is built on the Olmo-3 family, a fully open LLM family whose competitive performance aligns with state-of-the-art open LLMs.
Although \method\space offers a practical testbed, extending our evaluation protocol to other models and scales warrants future exploration.
Furthermore, while \method\space covers standard post-training pipelines (SFT, DPO, and RL), emerging paradigms like on-policy distillation can be incorporated in future work to further broaden our evaluations.

\clearpage




\bibliography{iclr2027_conference}
\bibliographystyle{iclr2027_conference}

\appendix
\clearpage

\addcontentsline{toc}{section}{Appendix}
\part{Appendix} 
\parttoc 

\section{Related Work}
\paragraph{Training data detection for LLMs.}
Training data detection for LLMs has been studied across both pre-training and post-training stages~\citep{shidetecting, zhang2024fine}.
Existing detection methods generally compute a membership score based on input sequences, including techniques such as \textit{Perplexity}~\citep{li2023estimating}, \textit{Entropy}~\citep{song2021systematic}, \textit{Zlib}~\citep{carlini2021extracting}, 
\textit{Lowercase}~\citep{carlini2021extracting},
\textit{Neighbor}~\citep{mattern2023membership}, \textit{\textsc{Min-K\%}}~\citep{shidetecting}, and \textit{\textsc{Min-K\%++}}~\citep{zhang2024min}.
Reference-based methods further calibrate these scores using reference models to achieve better detection performance~\citep{carlini2022membership,fu2024membership}. 
Recent studies have developed methods for particular post-training paradigms~\citep{fu2024membership, zhang2025detecting}.
Specifically, prior work detects fine-tuning samples by calibrating the model's predicted probability using a reference model trained on target-generated data~\citep{fu2024membership}.
Detection methods targeting RL training data exploit specific behavioral signals, such as entropy changes or nearest-neighbor edit distances, to infer membership in the reinforcement learning stage~\citep{tao2026detecting, zhang2026detecting}.
These studies generally aim to develop more effective detection methods to identify training data from pre-training or post-training stages.
Instead, our work systematically examines whether existing methods remain effective when evaluated throughout a realistic, large-scale post-training setting.

\paragraph{Evaluation of training data detection.}
Existing evaluations obtain membership labels either through post-hoc benchmark construction or through controlled small-scale model training. 
For example, post-hoc construction treats data published before a model's training cutoff as members and data published after the cutoff as non-members, such as WikiMIA ~\citep{shidetecting, yi2026membership}. 
MIMIR establishes a public evaluation testbed by utilizing target models trained on the publicly available Pile corpus, with members and non-members sampled from the dataset's training and test splits, respectively~\citep{duan2024membership}. 
However, recent work shows that blind baselines can distinguish members from non-members on existing benchmarks, suggesting that detection methods may exploit distributional differences rather than the true membership signal~\citep{das2024blind, duan2024membership}.
Evaluations in controlled training settings avoid data distribution shift but often focus on a single post-training stage and use small-scale training~\citep{fu2024membership, tao2026detecting}.
To address these limitations, our work introduces \method, a reliable benchmark designed to evaluate detection methods in realistic, large-scale post-training
settings.
Specifically, our benchmark constructs members and non-members from the same source distributions and validates the reliability of evaluation sets using blind baselines.
While CheckMIABench~\citep{wang2026checkmiabench} obtains pre-training data membership labels from intermediate checkpoints and inferred sample ordering during training, our benchmark conducts controlled training across SFT, DPO, and RL, producing verifiable and stage-specific membership labels throughout the post-training lifecycle.
We provide corresponding oracle models throughout the full post-training lifecycle (\textit{SFT} $\rightarrow$ \textit{DPO}$\rightarrow$ \textit{RL}), enabling evaluation of training data detection in realistic, large-scale post-training scenarios.

\section{Experimental Details}
\label{exe_details}
\subsection{Model training configurations}
\label{model_training}
Building on the fully open Olmo 3 model families~\citep{olmo2025olmo}, we conduct our main experiments using the Olmo 3 base model and train with three post-training paradigms---SFT, DPO, and GRPO. 
Detailed training configurations are shown in Table~\ref{tab:olmo3_7b_training}.

\paragraph{SFT}
We perform instruction tuning on \textsc{Olmo-3-1025-7B} base model using the Dolci-Instruct-SFT dataset.
SFT training is conducted for two epochs, with a learning rate of $8.0 \times 10^{-5}$ and a maximum sequence length of 32K.

\paragraph{DPO}
We train \textsc{Olmo-3-7B-Instruct-SFT} for one epoch on the Dolci-Instruct-DPO dataset to obtain \textsc{Oracle-DPO}. 
Training is conducted on eight GPUs with a batch size of 32 and a maximum sequence length of 16K. 
We use a learning rate of $1.0 \times 10^{-6}$ with linear decay, a warmup ratio of 0.1, and set the DPO coefficient $\beta$ to 5.

\paragraph{GRPO}
We train \textsc{Olmo-3-7B-Instruct-DPO} on the Dolci-Instruct-RL dataset to obtain \textsc{Oracle-RL}. 
We perform model training for 450 steps with four minibatches per update and a constant learning rate of $1.0 \times 10^{-6}$. 
In our experiments, we use a group size of 8, sampling temperature of 1.0, and clipping thresholds of 0.2 and 0.272 for lower and upper bounds, respectively.
The maximum sequence length is 8K tokens, with prompt and response lengths capped at 2,048 and 8,192 tokens, respectively.
Additionally, we track all prompts whose sampled responses achieve non-zero advantages during RL training. 
The members of the evaluation set are selected exclusively from these prompts.

\begin{table}[t]
\centering
\caption{Training hyperparameters for the SFT, DPO, and RL stages.}
\label{tab:olmo3_7b_training}
\small
\renewcommand{\arraystretch}{1.2}
\setlength{\tabcolsep}{5pt}
\begin{tabular*}{\linewidth}{@{\extracolsep{\fill}}lccc@{}}
\toprule
\textbf{Hyperparameter}
& \textbf{\textsc{Oracle-SFT}}
& \textbf{\textsc{Oracle-DPO}}
& \textbf{\textsc{Oracle-RL}} \\
\midrule
Epochs                       & 2                   & 1                    & --                  \\
Training steps               & --                  & --                   & 450                 \\
Learning rate                & $8.0 \times 10^{-5}$ & $1.0 \times 10^{-6}$ & $1.0 \times 10^{-6}$ \\
DPO $\beta$                  & --                  & 5                    & --                  \\
Batch size                   & 32                  & 32                  & 64                  \\
Minibatches                  & --                  & --                   & 4                   \\
\addlinespace[2pt]
Maximum sequence length      & 32K                 & 16K                  & 8K                  \\
Group size                   & --                  & --                   & 8                   \\
Sampling temperature         & --                  & --                   & 1.0                 \\
Clip lower                   & --                  & --                   & 0.2                 \\
Clip higher                  & --                  & --                   & 0.272               \\
\addlinespace[2pt]
Training GPUs                & 8                   & 8                   & 32                  \\
\bottomrule
\end{tabular*}
\end{table}

\subsection{Details of training datasets and evaluation sets}
\label{dataset}

\begin{table}[t]
\centering
\caption{Construction details of the post-training and detection datasets.}
\label{tab:training_data_construction}
\renewcommand{\arraystretch}{1.3}
\resizebox{\textwidth}{!}{%
\setlength{\tabcolsep}{0.8mm}%
\begin{tabular}{cccccccc}
\toprule

\textbf{Training Stage}
& \textbf{Base Model}
& \textbf{Dataset Source}
& \textbf{Train Size}
& \textbf{Target Model}
& \textbf{Evaluation Sets} 
& \textbf{Members} 
& \textbf{Non-members} \\
\midrule

SFT & Olmo-3-1025-7B & Dolci-Instruct-SFT & 2.15M & \textsc{Oracle-SFT} & Math, Code, IF, QA & $4 \times 250$ & $4 \times 250$\\ 
\midrule

DPO & Olmo-3-7B-Instruct-SFT & Dolci-Instruct-DPO & 260K & \textsc{Oracle-DPO} & Math, Code, IF, QA & $4 \times 250$ & $4 \times 250$\\ 

\midrule
RL & Olmo-3-7B-Instruct-DPO & Dolci-Instruct-RL & 170K & \textsc{Oracle-RL}& Math, Code, IF, QA & $4 \times 250$ & $4 \times 250$\\

\bottomrule
\end{tabular}%
}
\end{table}
Table~\ref{tab:training_data_construction} presents the datasets used for training and evaluation. 
We apply SFT to \textsc{Olmo-3-1025-7B} using 2.15M samples from Dolci-Instruct-SFT, yielding \textsc{Oracle-SFT}. 
For DPO, we train \textsc{Olmo-3-7B-Instruct-SFT} on 260K preference pairs from Dolci-Instruct-DPO to obtain \textsc{Oracle-DPO}. 
For RL, we train \textsc{Olmo-3-7B-Instruct-DPO} on 170K samples from Dolci-Instruct-RL, producing \textsc{Oracle-RL}.
For each stage, the benchmark comprises four tasks: Math, Coding, IF, and QA. 
Each task contains 250 training members and 250 held-out non-members, drawn from the same data distribution, yielding 2,000 evaluation examples per stage.

\subsection{Evaluation protocols}
\label{expe:evaluation_details}
In our main experiments, all reported results are averaged over three runs with different random seeds during inference. 
Detailed standard deviations across repeated runs are provided in Appendix~\ref{auc_std}.
In our experiments, we assume access to complete sample information during detection.
Specifically, we evaluate training data detection using the following protocol.

\paragraph{Input-based methods}
\textit{Input-based methods} compute membership scores based on the likelihood of input tokens. 
For SFT, we compute the score on each training instance using the complete question-response pair $(q,y)$. 
For DPO, where each training example consists of a preference triplet $(q,y^+,y^-)$, we compute the membership score on the chosen question-response pair $(q,y^+)$.

For RL, we compute the score on the question $q$, since RL does not rely on target responses.

\paragraph{Output-based methods}
\textit{Output-based methods} compute membership scores exclusively over generated tokens. 
For SFT, we remove the ground-truth response from the sample and generate a response conditioned on the question. 
For DPO, we generate a response using the preferred dialogue, excluding its last assistant turn as the prompt.
For RL, we use the sample question as the prompt to generate a response.
Unless otherwise specified, we set the maximum sequence length to 2,048 tokens. 

\subsection{Training data detection methods for LLMs}
\label{methods}
Here, we summarize common training data detection methods for LLMs examined in this work. 

\paragraph{Perplexity} \textit{Perplexity}~\citep{li2023estimating} is proposed to identify training data, based on the observation that members tend to have lower perplexity than non-members. Thus, the perplexity is calculated as:
\begin{equation}
\operatorname{Perplexity}(x)
= \exp\left\{
-\frac{1}{n}\sum_{i=1}^{n}
\log p_\theta\left(x_i \mid x_1,\ldots,x_{i-1}\right)
\right\},
\end{equation}
where \(x = \{x_1, x_2, \ldots, x_n\}\) is a sequence of tokens and
\(p_\theta(x_i \mid x_1, \ldots, x_{i-1})\) is the conditional
probability of \(x_i\) given the preceding tokens.

\noindent
\paragraph{Entropy}
\textit{Entropy}~\citep{song2021systematic} measures the model's average predictive uncertainty across tokens. The score is defined as the mean token-level entropy:
\begin{equation}
\mathrm{Entropy}(x)
=
-\frac{1}{n}
\sum_{i=1}^{n}
\sum_{v\in\mathcal{V}}
p_{\theta}(v\mid x_{<i})
\log p_{\theta}(v\mid x_{<i}),
\end{equation}
where $\mathcal{V}$ denotes the vocabulary.

\noindent
\paragraph{Lowercase}
\textit{Lowercase}~\citep{carlini2021extracting} normalizes text perplexity by the perplexity of the lowercase counterpart:
\begin{equation}
\mathrm{Lowercase}(x)
=
-\frac{
\operatorname{PPL}_{\theta}(x)
}{
\operatorname{PPL}_{\theta}(\operatorname{lower}(x))
},
\end{equation}
where $\operatorname{lower}(x)$ is the lowercase version of $x$.

\noindent
\paragraph{Zlib}
\textit{Zlib}~\citep{carlini2021extracting} uses the negative ratio of log-perplexity to the zlib-compressed length:
\begin{equation}
\mathrm{Zlib}(x)
=
-\frac{\log \operatorname{PPL}_{\theta}(x)}
{L_{\mathrm{zlib}}(x)},
\end{equation}
\noindent
where \(L_{\mathrm{zlib}}(x)\) denotes the size of the zlib-compressed representation of \(x\).

\noindent
\paragraph{Neighbor}
\textit{Neighbor}~\citep{mattern2023membership} perturbs the input sentence with masked language
models to create “neighbor” and compare the loss of the input sentence with the average loss of the neighbor sentences. The score is calculated as:
\begin{equation}
\mathrm{Neighbor}(x)
=
\frac{1}{m}
\sum_{j=1}^{m}
\mathcal{L}_{\theta}(\widetilde{x}_j)
-
\mathcal{L}_{\theta}(x),
\end{equation}
where $\{\widetilde{x}_j\}_{j=1}^{m}$ are $m$ neighboring texts and $\mathcal{L}_{\theta}(x)$ is the loss of the original text.

\noindent
\paragraph{Min-K\%}
\textit{Min-K\%}~\citep{shidetecting} computes the average log-probability of the $k\%$ tokens with the lowest predicted probabilities. Formally, \textsc{Min-K\%} is defined as:
\begin{equation}
\mathrm{Min\text{-}K\%}(x)
=
\frac{1}{E}
\sum_{x_i \in \mathrm{Min\text{-}K\%}(x)}
\log p_{\theta}
\left(x_i \mid x_1,\ldots,x_{i-1}\right),
\end{equation}
\noindent
where \(E\) denotes the number of tokens in \(\mathrm{Min\text{-}K\%}(x)\).

\noindent
\paragraph{Min-K\%++}
\textit{Min-K\% ++}~\citep{zhang2024min} normalizes the log-probability of each token using the mean and standard deviation of the predictive distribution over the vocabulary. 
The sequence-level score is obtained by averaging the lowest $k\%$ of normalized token scores:
\begin{equation}
\mathrm{Min\text{-}K\%++}(x)
=
\frac{1}{E}
\sum_{x_i \in \mathrm{Min\text{-}K\%++}(x)}
\frac{
\log p_{\theta}(x_i \mid x_{<i})-\mu_{x_{<i}}
}{
\sigma_{x_{<i}}
},
\end{equation}
\noindent
where \(E\) is the number of selected tokens, and
\begin{equation}
\mu_{x_{<i}}
=
\mathbb{E}_{z\sim p_{\theta}(\cdot\mid x_{<i})}
\left[
\log p_{\theta}(z\mid x_{<i})
\right],
\end{equation}
\begin{equation}
\sigma_{x_{<i}}
=
\sqrt{
\mathbb{E}_{z\sim p_{\theta}(\cdot\mid x_{<i})}
\left[
\left(
\log p_{\theta}(z\mid x_{<i})-\mu_{x_{<i}}
\right)^2
\right]
}.
\end{equation}

Following prior work, we set $k=20$ for both \textsc{Min-K\%} and \textsc{Min-K\%++} in all experiments.

\noindent
\paragraph{Min-kNN}
\textit{Min-kNN}~\citep{zhang2026detecting} samples multiple responses for an input and computes the average of the \(k\) smallest nearest-neighbor edit distances:
\begin{equation}
\mathrm{Min\text{-}kNN}(x)
=
\frac{1}{k}
\sum_{i=1}^{k} d_{(i)}(x),
\end{equation}
where \(d_{(i)}(x)\) denotes the \(i\)-th smallest nearest-neighbor normalized Levenshtein distance among the generated responses.

\paragraph{Self-Critique}
\textit{Self-Critique}~\citep{tao2026detecting} generates an initial response and a self-critique response for a given input, and then compares their token-level entropy sequences to measure RL-induced policy collapse.
The detection score is computed using a length-aware cosine similarity:
\begin{equation}
\mathrm{Self\text{-}Critique}(x)
=
\cos(\tilde{E}_1, \tilde{E}_2) \cdot
\frac{\min(|E_1|, |E_2|)}{\max(|E_1|, |E_2|)},
\end{equation}
where $E_1$ and $E_2$ denote the token-level entropy sequences of the two generated responses, and $\tilde{E}_1$ and $\tilde{E}_2$ are their zero-padded counterparts aligned to equal length.


\noindent
\paragraph{CDD}
\textit{CDD}~\citep{dong-etal-2024-generalization} detects training data based on the peakedness of the output distribution. The intuition is that, for a seen sample, the model tends to generate similar responses across multiple samplings, so the output distribution is sharply peaked. Given an input $x$, CDD first obtains a greedy response $y_0$ and then samples $n$ responses $\{y_1,\ldots,y_n\}$ with temperature sampling. The score is the fraction of sampled responses whose edit distance to $y_0$ is within a threshold:
\begin{equation}
\mathrm{CDD}(x)
=
\frac{1}{n}
\sum_{i=1}^{n}
\mathbb{I}\left(\mathrm{ED}(y_i, y_0) \le \alpha \cdot l\right),
\end{equation}
where $\mathrm{ED}(\cdot,\cdot)$ denotes the token-level edit distance, $l$ is the maximum length of the sampled responses, and $\alpha \in [0,1]$ controls the threshold. A higher score indicates a more peaked distribution and therefore a higher likelihood that $x$ is a training sample.

\noindent
\paragraph{Output-based Methods}
Given a prompt \(x\), we sample a response \(y\) from the target model. 
\textit{Generated Perplexity} computes the perplexity of the generated tokens, and \textit{Generated Entropy} measures the average predictive entropy of generated tokens. 
\textit{Generated Min-K\%} computes the average log-likelihood of K\% generated tokens with the lowest predicted probability.

\noindent
\paragraph{Reference-based Methods}
\textit{Reference-based methods}~\citep{carlini2021extracting} calibrate the membership score of a target model using the scores from a reference model:
\begin{equation}
S_{\mathrm{ref}}(x)
=
S(x;\mathcal{M})
-
S(x;\mathcal{M}_{\mathrm{ref}}),
\end{equation}
where \(S(x;\mathcal{M})\) and \(S(x;\mathcal{M}_{\mathrm{ref}})\) denote the metric scores of sample \(x\) computed by the target model \(\mathcal{M}\) and the reference model \(\mathcal{M}_{\mathrm{ref}}\), respectively.

\subsection{Implementation of small-scale training settings}
\label{expe_revisiting}
Our experiments in Section~\ref{main_result} show that existing detection methods exhibit limited performance under our large-scale post-training setting, making it difficult to clearly observe the effects of different evaluation settings.
We therefore adopt a small-scale training setting for the experimental analyses.
Specifically, we fine-tune the Olmo 3 base model on 10K examples sampled from Dolci-Instruct-SFT for 5 epochs, yielding \textsc{Oracle-SFT-10K}. 
By using a smaller training set and more training epochs, we amplify the membership signal and better examine the effects of different evaluation settings on detection performance.

\subsection{Implementation of threshold calibration}
\label{expe_threshold}
For threshold calibration on a validation set, we randomly sample 50 members and 50 non-members from the evaluation sets, then select the threshold that achieves the highest accuracy on the validation set. 
For ID calibration, the validation set is sampled from the Coding task.
For OOD calibration, our validation set is drawn from IF, QA, or Math tasks.
Each experiment is repeated 10 times with different random seeds, which produce different validation sets.
For threshold calibration methods without a validation set, we evaluate Gaussian Mixture Model (GMM), $\mathrm{GMM}_{\log1p}$, and BlindMI~\citep{BlindMI}.
GMM fits a two-component Gaussian mixture to the detection scores, while $\mathrm{GMM}_{\log1p}$ applies a $\log(1+x)$ transformation before fitting the same model. 
Without requiring a validation set, BlindMI initializes a potential non-member set from the test samples and iteratively refines it through differential comparison.

\subsection{Implementation of detection methods}
For CDD~\citep{dong-etal-2024-generalization}, only the prompt of each sample is used as input, and the ground-truth response is excluded. Based on this prompt, the target model generates one response via greedy decoding and $N$ responses via temperature sampling. 
All responses are truncated to their first 100 tokens before computing edit distances, with the threshold ratio $\alpha$ set to 0.05.

\subsection{Training dynamics and downstream performance of oracle models}
\begin{figure}[t]
    \centering
    \begin{subfigure}{0.48\linewidth}
        \centering
        \includegraphics[width=\linewidth]{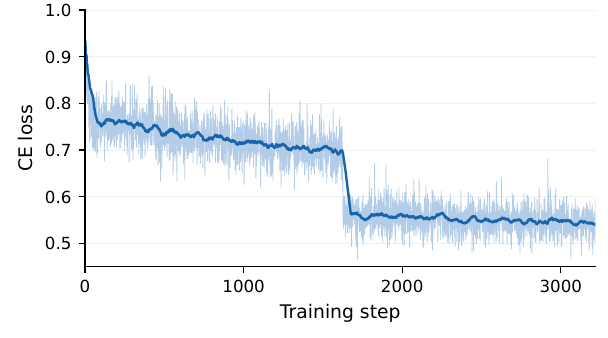}
        \subcaption{SFT cross-entropy loss}
        \label{fig:sft_loss}
    \end{subfigure}
    \hfill
    \begin{subfigure}{0.48\linewidth}
        \centering
        \includegraphics[width=\linewidth]{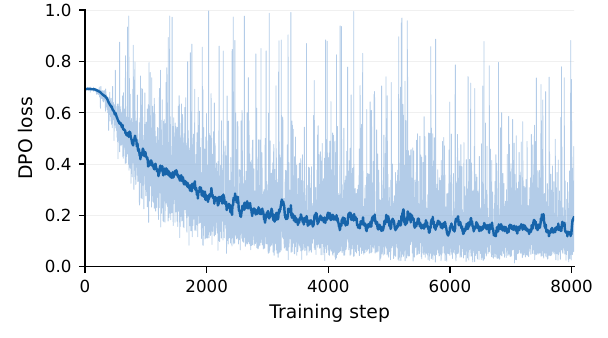}
        \subcaption{DPO preference loss}
        \label{fig:dpo_loss}
    \end{subfigure}
\caption{
    Training loss curves of the SFT and DPO post-training stages.
    (a) Cross-entropy loss during supervised fine-tuning (SFT).
    (b) Preference loss during direct preference optimization (DPO).
    Light lines show raw per-step losses, and dark lines show a
    50-step moving average.
}
    \label{fig:sft_dpo_loss}
\end{figure}

\begin{figure}[t]
    \centering
    \begin{subfigure}{0.48\linewidth}
        \centering
        \includegraphics[width=\linewidth]{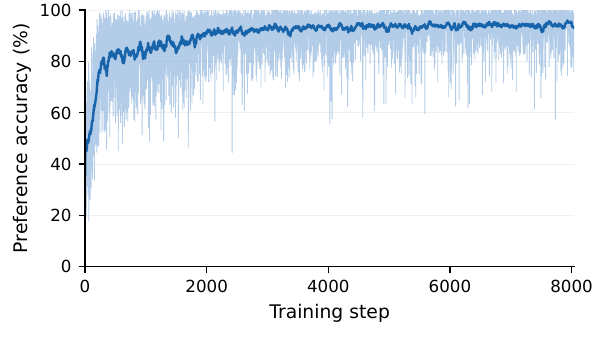}
        \subcaption{Preference accuracy}
        \label{fig:dpo_accuracy}
    \end{subfigure}
    \hfill
    \begin{subfigure}{0.48\linewidth}
        \centering
        \includegraphics[width=\linewidth]{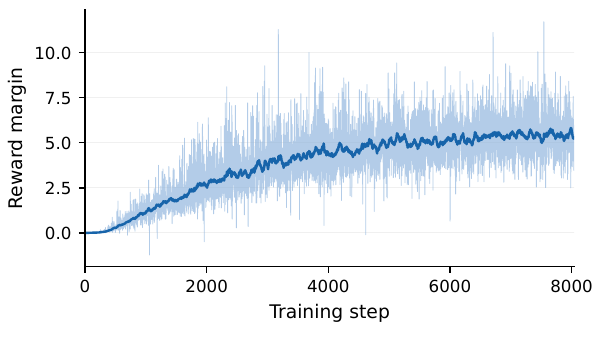}
        \subcaption{Reward margin}
        \label{fig:dpo_margin}
    \end{subfigure}
\caption{
    Training dynamics of the DPO post-training stage.
    (a) Preference accuracy: the fraction of preference pairs in which
    the chosen response receives a higher implicit reward than the
    rejected one.
    (b) Reward margin: the mean difference in implicit reward between
    chosen and rejected responses.
    Light lines show raw per-step values, and dark lines show a
    50-step moving average.
}
    \label{fig:dpo_training_dynamics}
\end{figure}

\begin{figure}[t]
    \centering
    \begin{subfigure}{0.48\linewidth}
        \centering
        \includegraphics[width=\linewidth]{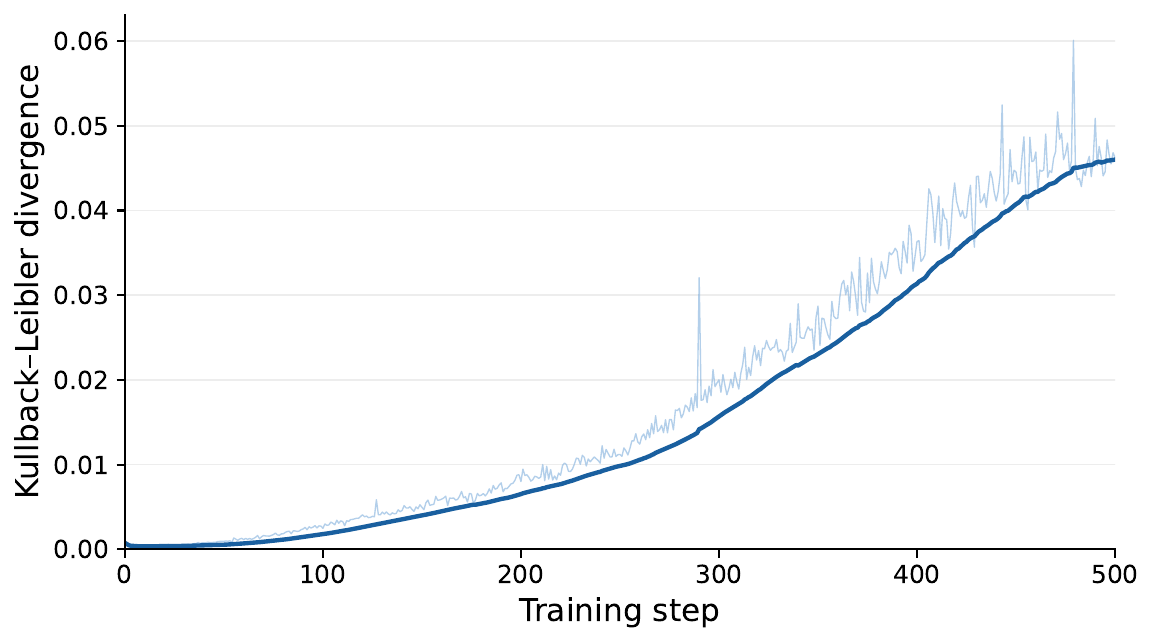}
        \subcaption{KL divergence}
        \label{fig:rl_kl}
    \end{subfigure}
    \hfill
    \begin{subfigure}{0.48\linewidth}
        \centering
        \includegraphics[width=\linewidth]{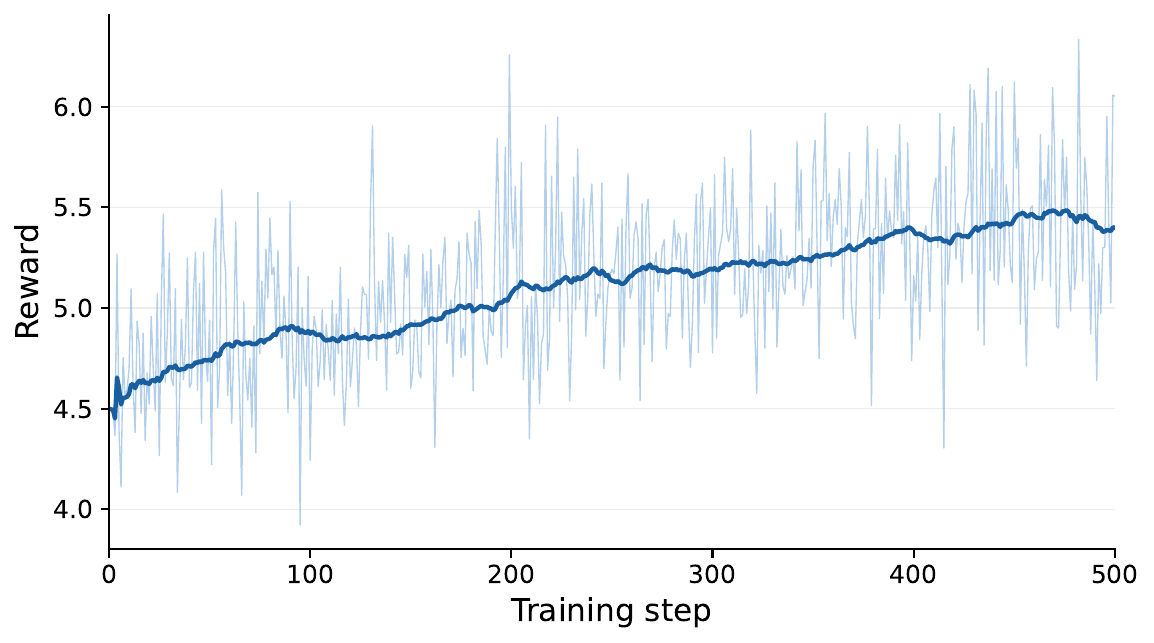}
        \subcaption{Reward}
        \label{fig:rl_reward}
    \end{subfigure}
\caption{
    Training dynamics of the RL post-training stage.
    (a) KL divergence between the policy and the reference model.
    (b) Mean reward of sampled responses at each training step.
    Light lines show raw per-step values, and dark lines show a
    50-step moving average.
}
    \label{fig:rl_dynamics}
\end{figure}

We report the training dynamics at each stage and the performance of the downstream task after each stage.
Figure~\ref{fig:sft_dpo_loss} shows the training loss of the SFT and DPO stages. Both losses decrease steadily and converge, indicating stable optimization. 
Figure~\ref{fig:dpo_training_dynamics} further shows that both preference accuracy and reward margin increase steadily during DPO, indicating that the model learns to prefer chosen responses over rejected ones.
Figure~\ref{fig:rl_dynamics} shows that the reward increases as training progresses, accompanied by a steady rise in KL divergence from the reference model.
This indicates that the policy is effectively optimized without training collapse.
We report the performance of oracle models on four downstream benchmarks in Table~\ref{tab:oracle_performance}. We conduct the evaluation at a temperature of 0.6 and a top-$p$ of 1.0, and report the pass@1 accuracy.
The results show that the accuracy of the downstream task improves across all four benchmarks as post-training progresses.
For example, accuracy on AIME increases from 3.33\% after SFT to 26.67\% after DPO and 40.00\% after RL. These results confirm that each post-training stage is well-trained, making our oracle models a reliable testbed for training data detection.
\begin{table}[t]
\centering
\footnotesize
\caption{
Downstream task performance (\%) of models after each post-training stage. The benchmark accuracy improves consistently from SFT to DPO to RL, confirming that each stage is effectively trained.
}
\label{tab:oracle_performance}
\renewcommand{\arraystretch}{1.2}
\setlength{\tabcolsep}{8mm}
\begin{tabular}{lccc}
\toprule
\textbf{Benchmark}
& \textsc{Oracle-SFT}
& \textsc{Oracle-DPO}
& \textsc{Oracle-RL} \\
\midrule
MATH500       & 60.00 & 77.80 & \textbf{89.80} \\
HumanEvalPlus & 63.41 & 68.90 & \textbf{76.83} \\
GPQA Diamond  & 31.82 & 38.89 & \textbf{41.41} \\
AIME 2024     &  3.33 & 26.67 & \textbf{40.00} \\
\bottomrule
\end{tabular}
\end{table}

\section{Additional Experimental Results}


\subsection{The composition of evaluation sets affects evaluation results}
\label{distribution_shift}
\begin{figure*}[t]
    \centering
    \begin{subfigure}{0.32\textwidth}
        \centering
        \includegraphics[width=\linewidth]{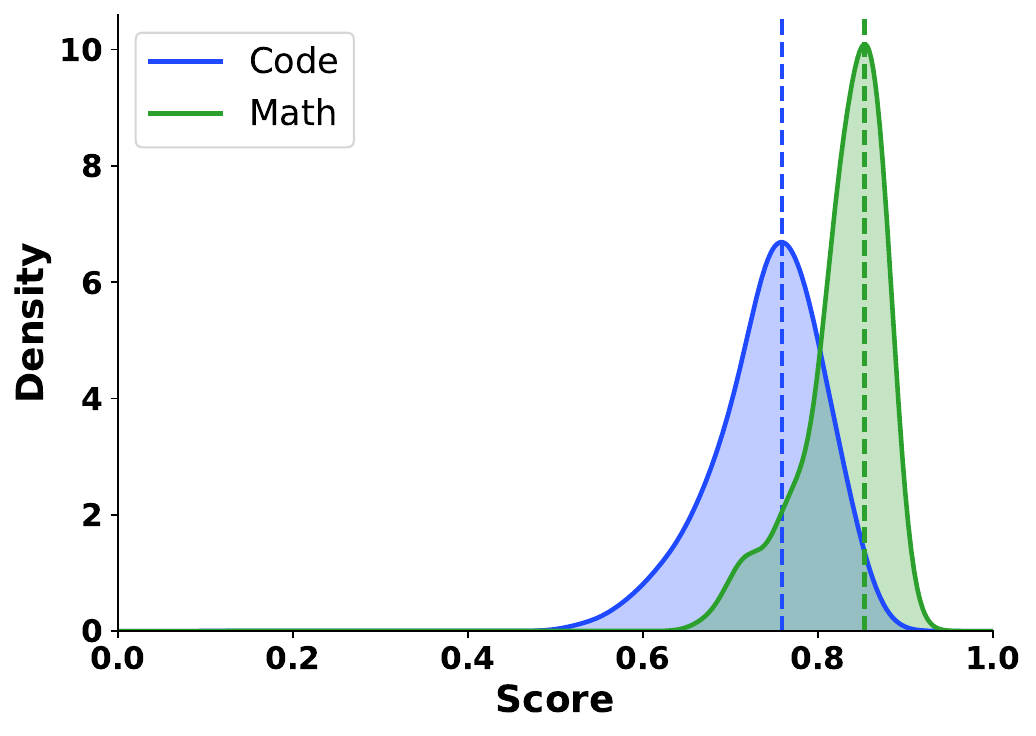}
        \caption{Code vs. Math (\textbf{AUC=0.879})}
        \label{fig:fir}
    \end{subfigure}
    \hfill
    \begin{subfigure}{0.32\textwidth}
        \centering
        \includegraphics[width=\linewidth]{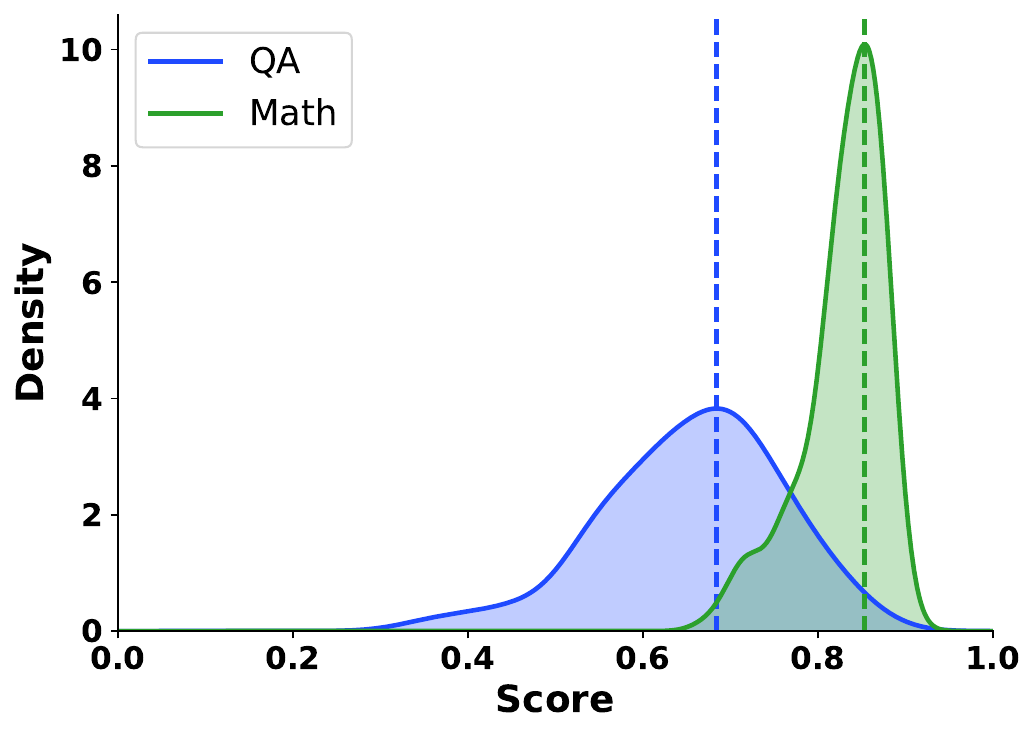}
        \caption{QA vs. Math (\textbf{AUC=0.949})}
        \label{fig:third}
    \end{subfigure}
    \hfill
    \begin{subfigure}{0.32\textwidth}
        \centering
        \includegraphics[width=\linewidth]{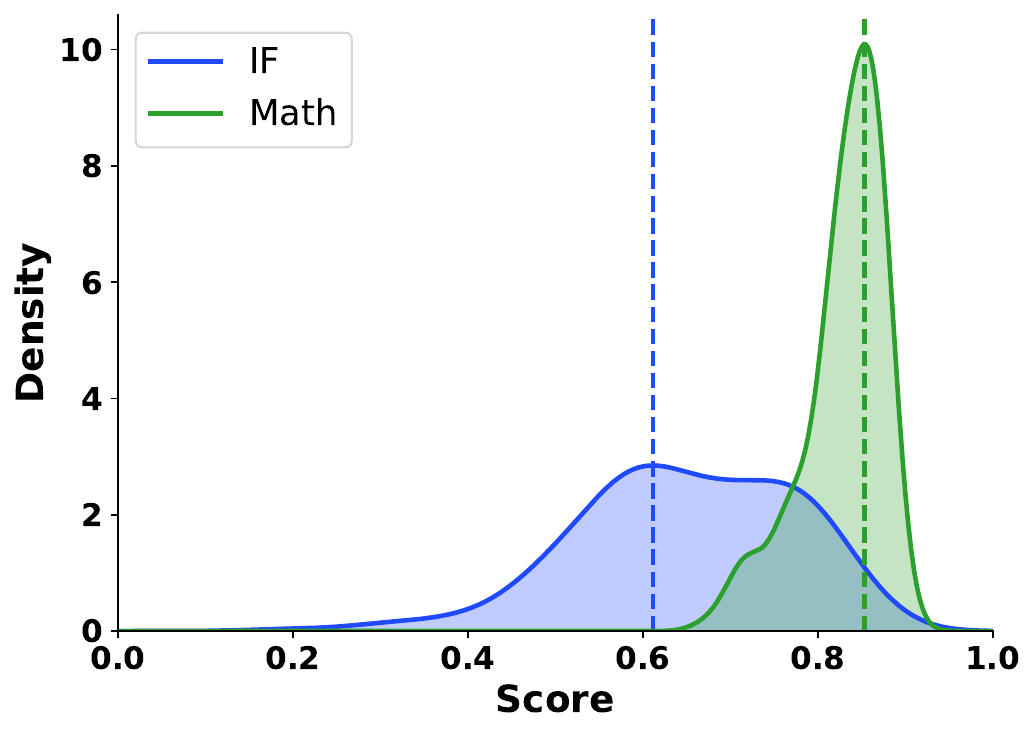}
        \caption{IF vs. Math (\textbf{AUC=0.929})}
        \label{fig:second}
\end{subfigure}
\caption{Score distributions of Math compared with QA, IF, and Code. Even when both groups consist entirely of non-members, domain distribution shifts can yield illusory detection performance.}
\label{fig:score_distribution}
\end{figure*}

Figure~\ref{fig:score_distribution} further illustrates the effect of domain distribution shifts on detection performance. 
Although both groups consist entirely of non-members, their membership score distributions differ substantially across domains.
We find that the Math task exhibits considerable distributional separation from other domains, even when both groups consist entirely of non-members. 
This separation is most pronounced with QA at an AUC of 0.95, moderate with IF at 0.93, and weakest but still notable with Coding at 0.88.
These results show that distribution shifts can lead to inflated AUC, even when both groups consist entirely of non-members. 

\subsection{Validating the reliability of our \method\space}
\label{no_shift}
\paragraph{Validating our \method\space with blind baselines.}
In Section~\ref{Validat_bench}, we validate the reliability of \method\space using blind baselines.
The results show that blind baselines obtain negligible signals for distinguishing members from non-members on our benchmark.
We provide additional implementation details and results here.
As described in Section~\ref{Validat_bench}, we construct nine blind baselines by combining three text feature extractors, Bag-of-Words, TF-IDF, and Word2Vec, with three classifiers, Logistic Regression, Random Forest, and Neural Network.
In our experiments, we conduct 10 runs with different random seeds for each attack baseline. 
Specifically, for each task, we randomly split the 250 members and 250 non-members into 400 training samples (200 members and 200 non-members) and 100 test samples (50 members and 50 non-members).
In Table~\ref{tab:feature_classifier_auc}, we report the mean and standard deviation of the AUC over the 10 runs.

\begin{figure*}[t]
    \centering
    \begin{subfigure}{0.43\textwidth}
        \centering
        \includegraphics[width=\linewidth]{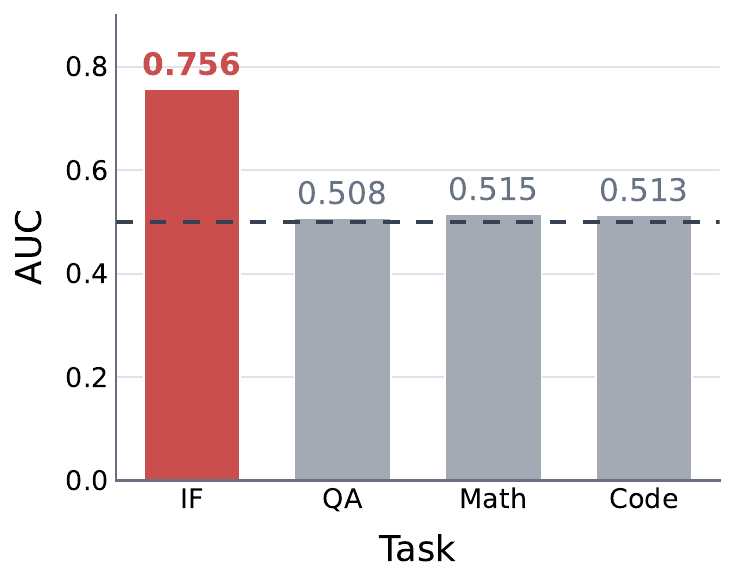}
        \caption{IF}
        \label{fig:score_dist_if}
    \end{subfigure}
    \hspace{1mm}
    \begin{subfigure}{0.43\textwidth}
        \centering
        \includegraphics[width=\linewidth]{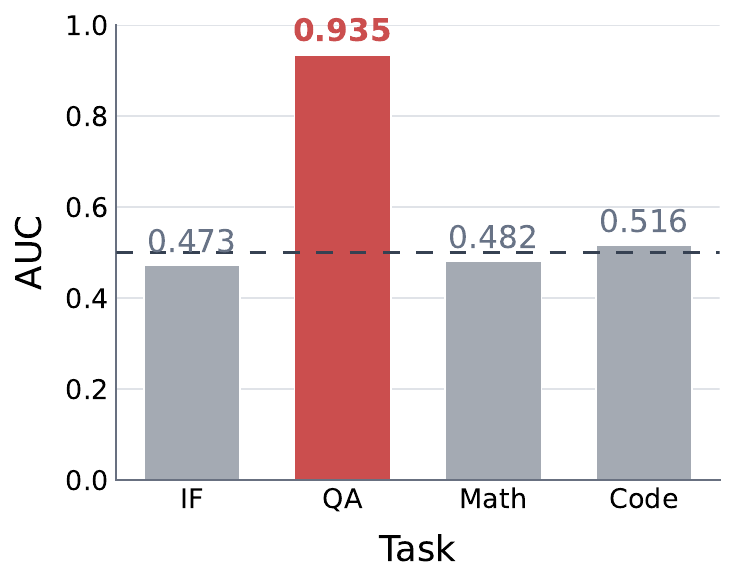}
        \caption{QA}
        \label{fig:score_dist_qa}
    \end{subfigure}
    \medskip
    \begin{subfigure}{0.43\textwidth}
        \centering
        \includegraphics[width=\linewidth]{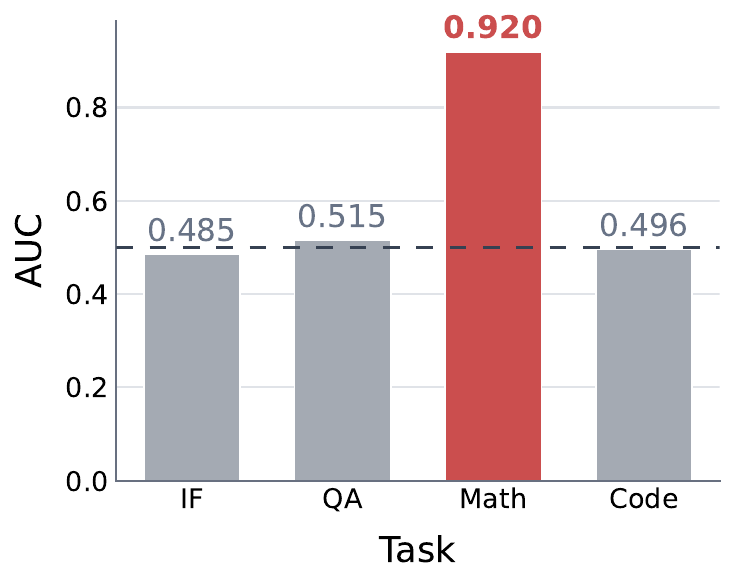}
        \caption{Math}
        \label{fig:score_dist_math}
    \end{subfigure}
    \hspace{1mm}
    \begin{subfigure}{0.43\textwidth}
        \centering
        \includegraphics[width=\linewidth]{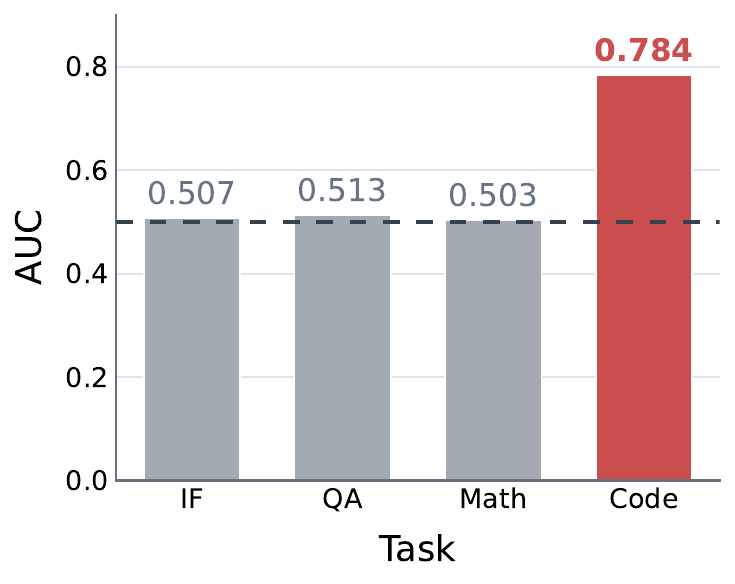}
        \caption{Coding}
        \label{fig:score_dist_Cod}
    \end{subfigure}
\caption{AUC of detection across fine-tuned models on a single-domain dataset. Models are fine-tuned separately on IF, QA, Math, and Coding tasks. Detection achieves a high AUC when evaluated on the trained task (red) but performs near a random guess when evaluated on untrained tasks (gray), confirming the reliability of our benchmark.}
\label{fig:auc_tasks}
\end{figure*}

\paragraph{Validating our \method\space with domain-specific models.}
To assess potential distributional shifts between the member and non-member subsets of our benchmark, we fine-tune the Olmo 3 base model separately on data from each of the four domains: IF, QA, Math, and Coding, yielding four domain-specific models.
We then evaluate each model on our evaluation sets. 
Figure~\ref{fig:auc_tasks} shows the AUC for detection across domain-specific models fine-tuned on a single-domain dataset.
The red bar represents evaluation on the training domain, where member samples are included in the training set.
The gray bars represent evaluation on unseen domains, for which evaluation sets are not included in the training.
Our results show that detection achieves high AUC when evaluated on the domain used for training, ranging from 0.756 to 0.935. 
In contrast, when evaluated on domains excluded from training, AUC ranges from 0.473 to 0.516, with an average of 0.502, close to random guessing.
These results show that a clear membership signal emerges only when evaluating on domains whose members were seen during training.
The near-chance AUC on the unseen domains suggests that differences between members and non-members are unlikely to yield effective detection performance.
These results suggest that the high detection performance on our benchmark comes from actual membership signals rather than exploitable distributional differences, thereby validating the reliability of our benchmark construction.

\begin{table*}[t]
\centering
\caption{
The AUC of blind baselines on our \method\space benchmark.
LR, RF, and NN denote Logistic Regression, Random Forest, and Neural Network, respectively. 
Results are reported as mean $\pm$ standard deviation over 10 random splits using 80\% of the data for training and 20\% for testing.
The best result for each task within each evaluation set is shown in \textbf{bold}.
}
\label{tab:feature_classifier_auc}

\renewcommand{\arraystretch}{1.3}
\setlength{\tabcolsep}{1mm}

\resizebox{\linewidth}{!}{%
\begin{tabular}{llccccccccc}
\toprule

\multirow{2}{*}{\textbf{Task}}
& \multirow{2}{*}{\textbf{Feature}}
& \multicolumn{3}{c}{\textbf{\textsc{Oracle-SFT}}}
& \multicolumn{3}{c}{\textbf{\textsc{Oracle-DPO}}}
& \multicolumn{3}{c}{\textbf{\textsc{Oracle-RL}}} \\

\cmidrule(lr){3-5}
\cmidrule(lr){6-8}
\cmidrule(lr){9-11}

&
& \textbf{LR} & \textbf{RF} & \textbf{NN}
& \textbf{LR} & \textbf{RF} & \textbf{NN}
& \textbf{LR} & \textbf{RF} & \textbf{NN} \\
\midrule

\multirow{3}{*}{Math}
& BoW
& $0.48 \pm 0.05$
& $0.49 \pm 0.04$
& $0.48 \pm 0.04$
& $0.48 \pm 0.06$
& $0.46 \pm 0.06$
& $0.50 \pm 0.07$
& $0.55 \pm 0.05$
& $0.55 \pm 0.02$
& $0.54 \pm 0.04$ \\

& TF-IDF
& $0.48 \pm 0.05$
& $0.47 \pm 0.03$
& $0.47 \pm 0.05$
& $0.50 \pm 0.03$
& $0.45 \pm 0.05$
& $0.51 \pm 0.03$
& $0.55 \pm 0.04$
& $0.54 \pm 0.06$
& $0.54 \pm 0.03$ \\

& Word2Vec
& $\mathbf{0.53 \pm 0.04}$
& $0.46 \pm 0.05$
& $0.45 \pm 0.04$
& $0.50 \pm 0.03$
& $0.49 \pm 0.04$
& $\mathbf{0.53 \pm 0.04}$
& $0.55 \pm 0.05$
& $0.53 \pm 0.04$
& $\mathbf{0.55 \pm 0.04}$ \\

\midrule

\multirow{3}{*}{Code}
& BoW
& $0.49 \pm 0.05$
& $0.51 \pm 0.04$
& $0.50 \pm 0.04$
& $0.51 \pm 0.02$
& $0.51 \pm 0.03$
& $0.49 \pm 0.01$
& $0.50 \pm 0.03$
& $\mathbf{0.52 \pm 0.05}$
& $0.48 \pm 0.03$ \\

& TF-IDF
& $0.51 \pm 0.06$
& $\mathbf{0.52 \pm 0.04}$
& $0.51 \pm 0.05$
& $0.51 \pm 0.01$
& $0.50 \pm 0.03$
& $0.50 \pm 0.02$
& $0.49 \pm 0.04$
& $0.49 \pm 0.04$
& $0.47 \pm 0.04$ \\

& Word2Vec
& $0.48 \pm 0.03$
& $0.51 \pm 0.06$
& $0.46 \pm 0.04$
& $\mathbf{0.51 \pm 0.06}$
& $0.51 \pm 0.04$
& $0.50 \pm 0.03$
& $0.50 \pm 0.04$
& $0.48 \pm 0.05$
& $0.48 \pm 0.05$ \\

\midrule

\multirow{3}{*}{IF}
& BoW
& $0.54 \pm 0.05$
& $0.52 \pm 0.06$
& $0.52 \pm 0.05$
& $0.54 \pm 0.05$
& $0.49 \pm 0.05$
& $0.54 \pm 0.03$
& $0.47 \pm 0.05$
& $0.50 \pm 0.04$
& $0.45 \pm 0.03$ \\

& TF-IDF
& $0.51 \pm 0.06$
& $0.54 \pm 0.04$
& $0.49 \pm 0.06$
& $0.53 \pm 0.04$
& $0.51 \pm 0.03$
& $0.54 \pm 0.03$
& $0.48 \pm 0.03$
& $0.51 \pm 0.03$
& $0.46 \pm 0.04$ \\

& Word2Vec
& $0.52 \pm 0.03$
& $0.56 \pm 0.04$
& $\mathbf{0.56 \pm 0.02}$
& $0.54 \pm 0.04$
& $\mathbf{0.56 \pm 0.04}$
& $0.56 \pm 0.04$
& $\mathbf{0.52 \pm 0.04}$
& $0.46 \pm 0.03$
& $0.49 \pm 0.04$ \\

\midrule

\multirow{3}{*}{QA}
& BoW
& $0.47 \pm 0.03$
& $0.47 \pm 0.04$
& $0.47 \pm 0.04$
& $0.51 \pm 0.04$
& $0.51 \pm 0.06$
& $0.50 \pm 0.03$
& $0.46 \pm 0.04$
& $0.45 \pm 0.03$
& $0.48 \pm 0.05$ \\

& TF-IDF
& $0.50 \pm 0.03$
& $0.46 \pm 0.05$
& $0.49 \pm 0.03$
& $0.51 \pm 0.04$
& $0.48 \pm 0.05$
& $\mathbf{0.51 \pm 0.05}$
& $0.49 \pm 0.05$
& $0.47 \pm 0.05$
& $\mathbf{0.50 \pm 0.05}$ \\

& Word2Vec
& $\mathbf{0.52 \pm 0.03}$
& $0.50 \pm 0.04$
& $0.48 \pm 0.03$
& $0.48 \pm 0.04$
& $0.47 \pm 0.05$
& $0.48 \pm 0.05$
& $0.50 \pm 0.03$
& $0.47 \pm 0.05$
& $0.47 \pm 0.03$ \\

\bottomrule
\end{tabular}%
}
\end{table*}

\subsection{Standard deviation of repeated experiments}
\label{auc_std}
To verify that reported metrics in our experiments are not driven by stochastic decoding, we repeat each experiment using three different inference random seeds.
We report the standard deviation of AUC in Table~\ref{tab:main_auc_std}. 
Most input-based methods are deterministic and yield zero variance, whereas Neighbor exhibits minor fluctuations due to its stochastic neighbor sampling procedure.
Output-based methods generally exhibit negligible standard deviations across runs, with Self-Critique displaying mild sensitivity under specific settings.
Overall, these results confirm that our findings are robust to inference randomness and are not an artifact of stochastic decoding.

\begin{table*}[!t]
\centering
\footnotesize
\caption{Standard deviation of AUC over three runs with different inference random seeds.
\textit{Input-based methods} indicate methods that compute a score using input tokens, whereas \textit{Output-based methods} score output tokens.}
\label{tab:main_auc_std}
\renewcommand{\arraystretch}{1.2}
\resizebox{\textwidth}{!}{%
\setlength{\tabcolsep}{1.5mm}%
\begin{tabular}{@{}lccccccccccccc@{}}
\toprule

\multirow{2}{*}{\textbf{Method}}
& \multicolumn{4}{c}{\textbf{\textsc{Oracle-SFT}}}
& \multicolumn{4}{c}{\textbf{\textsc{Oracle-DPO}}}
& \multicolumn{4}{c}{\textbf{\textsc{Oracle-RL}}}
& \multirow{2}{*}{\textbf{Avg.}} \\

\cmidrule(lr){2-5}
\cmidrule(lr){6-9}
\cmidrule(lr){10-13}

& Math & Code & IF & QA
& Math & Code & IF & QA
& Math & Code & IF & QA
& \\

\midrule

\rowcolor{gray!8}
\multicolumn{14}{l}{
    \textcolor{gray}{\textbf{\textit{Input-based methods}}}
} \\

\addlinespace[2pt]

Perplexity
& 0.000 & 0.000 & 0.000 & 0.000
& 0.000 & 0.000 & 0.000 & 0.000
& 0.000 & 0.000 & 0.000 & 0.000
& 0.000 \\

Entropy
& 0.000 & 0.000 & 0.000 & 0.000
& 0.000 & 0.000 & 0.000 & 0.000
& 0.000 & 0.000 & 0.000 & 0.000
& 0.000 \\

Lowercase
& 0.000 & 0.000 & 0.000 & 0.000
& 0.000 & 0.000 & 0.000 & 0.000
& 0.000 & 0.000 & 0.000 & 0.000
& 0.000 \\

Zlib
& 0.000 & 0.000 & 0.000 & 0.000
& 0.000 & 0.000 & 0.000 & 0.000
& 0.000 & 0.000 & 0.000 & 0.000
& 0.000 \\

Neighbor
& 0.007 & 0.006 & 0.003 & 0.000
& 0.012 & 0.009 & 0.008 & 0.003
& 0.006 & 0.008 & 0.007 & 0.002
& 0.006 \\

\textsc{MIN-K\%}
& 0.000 & 0.000 & 0.000 & 0.000
& 0.000 & 0.000 & 0.000 & 0.000
& 0.000 & 0.000 & 0.000 & 0.000
& 0.000 \\

\textsc{MIN-K\%++}
& 0.000 & 0.000 & 0.000 & 0.000
& 0.000 & 0.000 & 0.000 & 0.000
& 0.000 & 0.000 & 0.000 & 0.000
& 0.000 \\

\addlinespace[3pt]

\rowcolor{gray!8}
\multicolumn{14}{l}{
    \textcolor{gray}{\textbf{\textit{Output-based methods}}}
} \\

\addlinespace[2pt]

CDD
& 0.001 & 0.005 & 0.004 & 0.003
& 0.007 & 0.010 & 0.005 & 0.008
& 0.007 & 0.006 & 0.009 & 0.004
& 0.006 \\

Generated Perplexity
& 0.004 & 0.001 & 0.001 & 0.002
& 0.004 & 0.003 & 0.001 & 0.001
& 0.006 & 0.003 & 0.003 & 0.001
& 0.003 \\

Generated Entropy
& 0.003 & 0.001 & 0.002 & 0.002
& 0.004 & 0.004 & 0.001 & 0.000
& 0.006 & 0.002 & 0.005 & 0.002
& 0.003 \\

Generated \textsc{MIN-K\%}
& 0.004 & 0.001 & 0.001 & 0.004
& 0.004 & 0.003 & 0.002 & 0.001
& 0.005 & 0.004 & 0.002 & 0.001
& 0.003 \\

Min-kNN
& 0.001 & 0.002 & 0.000 & 0.001
& 0.001 & 0.001 & 0.002 & 0.000
& 0.005 & 0.002 & 0.001 & 0.003
& 0.002 \\

Self-Critique
& 0.002 & 0.004 & 0.006 & 0.014
& 0.020 & 0.094 & 0.046 & 0.046
& 0.010 & 0.006 & 0.009 & 0.008
& 0.022 \\

\bottomrule
\end{tabular}%
}
\end{table*}

\begingroup

\subsection{Evaluating DPO-specific methods on our benchmark}
\begin{table*}[!t]
\centering
\footnotesize
\caption{AUC and TPR@5\%FPR of PREMIA with different reference models
on WildTrace-DPO. \textit{PREMIA-Base} uses OLMo-3-1025-7B,
whereas \textit{PREMIA-SFT} uses OLMo-3-7B-Instruct-SFT
as the reference model. Avg. denotes the average across four domains.
\textbf{Bold} indicates the superior result.}
\label{tab:premia_auc}
\renewcommand{\arraystretch}{1.2}

\resizebox{\textwidth}{!}{%
\setlength{\tabcolsep}{3mm}%
\begin{tabular}{@{}lcccccccccc@{}}
\toprule

\multirow{2}{*}{\textbf{Method}}
& \multicolumn{5}{c}{\textbf{AUC}}
& \multicolumn{5}{c}{\textbf{TPR@5\%FPR}} \\

\cmidrule(lr){2-6}
\cmidrule(lr){7-11}

& Math & Code & IF & QA & Avg.
& Math & Code & IF & QA & Avg. \\

\midrule

PREMIA-Base
& 0.517 & \textbf{0.474} & \textbf{0.537}
& \textbf{0.572} & \textbf{0.525}
& 0.048 & \textbf{0.056} & 0.048
& 0.052 & 0.051 \\

PREMIA-SFT
& \textbf{0.519} & 0.472 & 0.528
& 0.571 & 0.522
& \textbf{0.052} & 0.028 & \textbf{0.072}
& \textbf{0.100} & \textbf{0.063} \\

\bottomrule
\end{tabular}%
}
\end{table*}
To examine whether preference-aware signals improve the detection of DPO training data, we evaluate PREMIA~\citep{feng2024exposing}, a detection method designed for preference data. 
Specifically, PREMIA computes the probability ratio of each response between the target model and a reference model, and uses the difference between the ratios of the chosen and rejected responses as the membership score.
In our experiments, we consider two reference models: the base model OLMo-3-1025-7B (PREMIA-Base) and the preceding checkpoint OLMo-3-7B-Instruct-SFT (PREMIA-SFT). As shown in Table~\ref{tab:premia_auc}, PREMIA achieves limited detection performance on \textsc{Oracle-DPO} with both reference models, with an average AUC of at most 0.525 and TPR@5\%FPR of at most 0.063. 
These results indicate that even methods tailored to preference data exhibit limited effectiveness under large-scale DPO training. 

\subsection{Evaluating reference-based methods on oracle models}
\label{sec:reference-based}
Table~\ref{tab:reference_auc} reports the AUC of reference-based methods on the three oracle models. For each oracle model, we use the checkpoint before the corresponding post-training stage as the reference model.
The results show that reference-based methods also exhibit limited detection performance, with average AUCs ranging from 0.504 to 0.604.
This suggests that introducing a reference model yields limited improvement in detecting training data under large-scale post-training.
\begin{table*}[!t]
\centering
\footnotesize
\caption{Comparison of detection performance with and without reference models on our benchmark. We use OLMo-3-1025-7B, OLMo-3-7B-Instruct-SFT, and OLMo-3-7B-Instruct-DPO as reference models for \textsc{Oracle-SFT}, \textsc{Oracle-DPO}, and \textsc{Oracle-RL}, respectively. Avg. denotes the mean AUC across all tasks and post-trained models. $\Delta$ denotes the change in Avg. after introducing reference models. \textbf{Bold} highlights the higher AUC for each method across the two settings.}
\label{tab:reference_auc}
\renewcommand{\arraystretch}{1.2}

\resizebox{\textwidth}{!}{%
\setlength{\tabcolsep}{1.5mm}%
\begin{tabular}{@{}ll*{14}{c}@{}}
\toprule

\multirow{3}{*}{\textbf{Method}}
& \multirow{3}{*}{\textbf{Setting}}
& \multicolumn{4}{c}{\textbf{\textsc{Oracle-SFT}}}
& \multicolumn{4}{c}{\textbf{\textsc{Oracle-DPO}}}
& \multicolumn{4}{c}{\textbf{\textsc{Oracle-RL}}}
& \multirow{3}{*}{\textbf{Avg.}}
& \multirow{3}{*}{$\boldsymbol{\Delta}$} \\

& 
& \multicolumn{4}{c}{\scriptsize OLMo-3-1025-7B}
& \multicolumn{4}{c}{\scriptsize OLMo-3-7B-Instruct-SFT}
& \multicolumn{4}{c}{\scriptsize OLMo-3-7B-Instruct-DPO}
& & \\

\cmidrule(lr){3-6}
\cmidrule(lr){7-10}
\cmidrule(lr){11-14}

&
& Math & Code & IF & QA
& Math & Code & IF & QA
& Math & Code & IF & QA
& & \\

\midrule

\multirow{2}{*}{Perplexity}
& w/o Ref.
& 0.537 & 0.570 & 0.620 & 0.660
& 0.500 & 0.451 & 0.527 & 0.514
& 0.530 & \textbf{0.492} & 0.503 & \textbf{0.475}
& 0.532 & -- \\

& w/ Ref.
& \textbf{0.713} & \textbf{0.708} & \textbf{0.658} & \textbf{0.739}
& \textbf{0.503} & \textbf{0.467} & \textbf{0.531} & \textbf{0.546}
& \textbf{0.533} & 0.484 & \textbf{0.528} & 0.467
& \textbf{0.573} & \textcolor{red}{+0.042} \\

\midrule

\multirow{2}{*}{Entropy}
& w/o Ref.
& 0.501 & 0.520 & \textbf{0.486} & \textbf{0.444}
& \textbf{0.524} & 0.447 & 0.487 & 0.488
& \textbf{0.515} & 0.492 & \textbf{0.510} & 0.474
& 0.491 & -- \\

& w/ Ref.
& \textbf{0.555} & \textbf{0.572} & 0.435 & 0.401
& 0.508 & \textbf{0.521} & \textbf{0.511} & \textbf{0.492}
& 0.500 & \textbf{0.565} & 0.493 & \textbf{0.494}
& \textbf{0.504} & \textcolor{red}{+0.013} \\

\midrule

\multirow{2}{*}{Lowercase}
& w/o Ref.
& 0.595 & 0.593 & 0.628 & 0.673
& \textbf{0.498} & 0.438 & 0.514 & \textbf{0.523}
& 0.504 & 0.492 & \textbf{0.522} & 0.467
& 0.537 & -- \\

& w/ Ref.
& \textbf{0.722} & \textbf{0.642} & \textbf{0.682} & \textbf{0.746}
& 0.488 & \textbf{0.513} & \textbf{0.533} & 0.518
& \textbf{0.533} & \textbf{0.508} & 0.502 & \textbf{0.511}
& \textbf{0.575} & \textcolor{red}{+0.038} \\

\midrule

\multirow{2}{*}{Zlib}
& w/o Ref.
& 0.518 & 0.521 & 0.579 & 0.567
& 0.466 & \textbf{0.538} & 0.526 & 0.465
& 0.513 & \textbf{0.518} & 0.513 & \textbf{0.505}
& 0.519 & -- \\

& w/ Ref.
& \textbf{0.702} & \textbf{0.710} & \textbf{0.703} & \textbf{0.749}
& \textbf{0.495} & 0.532 & \textbf{0.542} & \textbf{0.519}
& \textbf{0.537} & 0.487 & \textbf{0.523} & 0.462
& \textbf{0.580} & \textcolor{red}{+0.061} \\

\midrule

\multirow{2}{*}{\textsc{Min-K\%}}
& w/o Ref.
& 0.540 & 0.569 & 0.606 & 0.662
& 0.496 & 0.459 & \textbf{0.535} & 0.519
& 0.526 & \textbf{0.491} & 0.502 & \textbf{0.476}
& 0.532 & -- \\

& w/ Ref.
& \textbf{0.777} & \textbf{0.765} & \textbf{0.795} & \textbf{0.848}
& \textbf{0.503} & \textbf{0.496} & 0.524 & \textbf{0.551}
& \textbf{0.552} & 0.463 & \textbf{0.514} & 0.456
& \textbf{0.604} & \textcolor{red}{+0.072} \\

\midrule

\multirow{2}{*}{\textsc{Min-K\%++}}
& w/o Ref.
& 0.602 & 0.615 & 0.646 & 0.758
& 0.504 & \textbf{0.543} & 0.473 & 0.544
& 0.545 & 0.462 & 0.508 & \textbf{0.491}
& 0.558 & -- \\

& w/ Ref.
& \textbf{0.695} & \textbf{0.670} & \textbf{0.684} & \textbf{0.810}
& \textbf{0.521} & 0.532 & \textbf{0.484} & \textbf{0.570}
& \textbf{0.552} & \textbf{0.488} & \textbf{0.542} & 0.488
& \textbf{0.586} & \textcolor{red}{+0.029} \\

\bottomrule
\end{tabular}%
}
\end{table*}

\subsection{Detectability of SFT data across post-training stages}
\begin{table*}[!t]
\centering

\caption{AUC and TPR@5\%FPR in detecting the same SFT training data on checkpoints from successive post-training stages. SFT denotes the model after SFT, SFT $\to$ DPO the model after further DPO training, and SFT $\to$ DPO $\to$ RL the model after further RL training. Avg. denotes the mean across the four domains. \textbf{Bold} indicates the best result in each column.}
\label{tab:training_stage_auc}
\small
\renewcommand{\arraystretch}{1.2}
\setlength{\tabcolsep}{2mm}
\begin{tabular}{@{}l ccccc ccccc@{}}
\toprule
\multirow{2}{*}{\textbf{Post-training Stages}}
  & \multicolumn{5}{c}{\textbf{AUC}}
  & \multicolumn{5}{c}{\textbf{TPR@5\%FPR}} \\
\cmidrule(lr){2-6} \cmidrule(l){7-11}
  & Math & Code & IF & QA & Avg.
  & Math & Code & IF & QA & Avg. \\
\midrule
SFT
  & \textbf{0.882} & \textbf{0.833} & \textbf{0.817} & \textbf{0.895} & \textbf{0.857}
  & \textbf{0.648} & \textbf{0.480} & \textbf{0.560} & \textbf{0.704} & \textbf{0.598} \\
SFT $\to$ DPO
  & 0.879 & 0.830 & 0.808 & 0.880 & 0.849
  & 0.644 & 0.456 & 0.556 & 0.684 & 0.585 \\
SFT $\to$ DPO $\to$ RL
  & 0.871 & 0.824 & 0.803 & 0.878 & 0.844
  & 0.644 & 0.436 & 0.552 & 0.688 & 0.580 \\
\bottomrule
\end{tabular}
\end{table*}
To examine how post-training affects earlier membership signals, we track the same SFT members and non-members across different post-training stages. 
To clearly observe how membership signals change across stages, we conduct this experiment under the small-scale setting in Section~\ref{sec:Revisiting}, where SFT samples exhibit clear membership signals. Specifically, starting from \textsc{Oracle-SFT-10K}, we sequentially perform DPO and RL training, each on 10K examples. 
We then evaluate the detection performance of the \textsc{Min-K\%++} method on WildTrace-SFT using the checkpoint obtained after each stage.
As shown in Table~\ref{tab:training_stage_auc}, subsequent DPO and RL training gradually weaken the detectability of SFT samples, with the average AUC decreasing from 0.857 to 0.844 and TPR@5\%FPR from 0.598 to 0.580. Nevertheless, most membership signals introduced during SFT persist through later post-training stages.
This suggests that although subsequent post-training weakens the membership signals in SFT data, they remain largely detectable in later stages.

\subsection{Effect of training data scale on detection performance}
\begin{table}[t]
\centering
\footnotesize
\caption{AUC of training data detection methods on \textsc{Oracle}-SFT models fine-tuned with different amounts of SFT data (10K to 1000K samples). We report the average AUC across four tasks. Detection performance generally degrades as the size of the SFT data increases.}
\label{tab:training_scale_all}
\renewcommand{\arraystretch}{1.15}
\setlength{\tabcolsep}{2mm}
\resizebox{0.9\textwidth}{!}{%
\begin{tabular}{lcccccc}
\toprule
\textbf{Model}
& \textbf{Perplexity}
& \textbf{Entropy}
& \textbf{Lowercase}
& \textbf{Zlib}
& \textbf{MIN-K\%}
& \textbf{MIN-K\%++} \\
\midrule
\textsc{Oracle-SFT-10K}
& 0.789 & 0.922 & 0.768 & 0.676 & 0.748 & 0.857 \\
\textsc{Oracle-SFT-50K}
& 0.786 & 0.929 & 0.776 & 0.674 & 0.748 & 0.872 \\
\textsc{Oracle-SFT-100K}
& 0.772 & 0.910 & 0.771 & 0.661 & 0.731 & 0.888 \\
\textsc{Oracle-SFT-200K}
& 0.761 & 0.905 & 0.761 & 0.652 & 0.728 & 0.868 \\
\textsc{Oracle-SFT-400K}
& 0.662 & 0.598 & 0.676 & 0.581 & 0.621 & 0.743 \\
\textsc{Oracle-SFT-1000K}
& 0.618 & 0.517 & 0.638 & 0.555 & 0.581 & 0.682 \\
\bottomrule
\end{tabular}%
}
\end{table}
To examine how training data scale affects detection performance, we evaluate various detection methods on models fine-tuned with different amounts of SFT data.
Specifically, we fine-tune the base model on 10K to 1000K SFT examples with all other configurations fixed, where each training set includes all benchmark members.
As shown in Table~\ref{tab:training_scale_all}, all methods exhibit a general decline as the training scale increases. 
For example, the AUC of Perplexity decreases from 0.789 to 0.618, and that of Entropy decreases from 0.922 to 0.517. 
These results confirm that strong detection performance observed in small-scale settings does not transfer to large-scale training across the evaluated methods.

\subsection{Evaluation on another model family}
\begin{table}[t]
\centering
\footnotesize
\caption{AUC of training data detection methods on the Apertus model family. Avg.\ denotes the mean AUC across the two datasets for each model. \textbf{Bold} indicates the best result.}
\label{tab:apertus}
\renewcommand{\arraystretch}{1.15}
\setlength{\tabcolsep}{5mm}
\begin{tabular}{lcccccc}
\toprule
\multirow{2}{*}{\textbf{Method}}
& \multicolumn{3}{c}{\textbf{Apertus-v1.5-8B}}
& \multicolumn{3}{c}{\textbf{Apertus-v1.5-70B-SFT}} \\
\cmidrule(lr){2-4}
\cmidrule(lr){5-7}
& Math & QA & Avg.
& GSM8K & QA & Avg. \\
\midrule
Perplexity        & 0.497 & 0.567 & \textbf{0.532} & 0.508 & 0.561 & \textbf{0.534} \\
Entropy           & \textbf{0.531} & 0.490 & 0.511 & 0.522 & 0.528 & 0.525 \\
Lowercase         & 0.491 & 0.497 & 0.494 & \textbf{0.524} & 0.479 & 0.501 \\
Zlib              & 0.484 & 0.548 & 0.516 & 0.490 & 0.545 & 0.517 \\
\textsc{Min-K\%}  & 0.488 & \textbf{0.570} & 0.529 & 0.501 & \textbf{0.563} & 0.532 \\
\textsc{Min-K\%++}& 0.468 & 0.535 & 0.502 & 0.490 & 0.549 & 0.519 \\
\bottomrule
\end{tabular}
\end{table}
To verify that our findings are not specific to OLMo-3, we conduct experiments on Apertus models~\citep{apertus2026apertus15}, a fully open model family at the 8B and 70B scales that publicly releases its training data. Due to limited computational resources, we construct the evaluation sets post hoc using publicly released training data. By examining the released post-training data, we find that the GSM8K and FeTaQA training splits are included in the model's training data. We therefore adopt GSM8K and FeTaQA for Math and QA tasks, respectively. For each dataset, we randomly sample 1,000 instances from the training split as members and 1,000 instances from the test split as non-members, yielding a balanced evaluation set.
Although post-hoc construction is not advocated, as discussed in Section~\ref{sec:Revisiting}, existing methods still exhibit limited detection performance on both 8B and 70B models, as shown in Table~\ref{tab:apertus}. 
These results suggest that our conclusions extend to larger models and other model families.
\endgroup

\subsection{Additional results}
\label{add_results}
Table~\ref{tab:main_tpr} shows the TPR@5\%FPR of various detection methods on our benchmark.
The experimental results show that existing methods are limited in effectiveness in large-scale post-training settings. 
Table~\ref{tab:domain_shift} shows the AUC of various detection methods across evaluation sets constructed with Math members and non-members from Math, Code, IF, or QA.
The results show that the detection method achieves
inflated AUC scores on the Math task when non-members are drawn from cross-domain sources.
\label{tpr}
\begin{table*}[!t]
\centering
\footnotesize
\caption{TPR@5\%FPR of detection methods on our benchmark. \textit{Input-based methods} indicate methods that compute a score using input tokens, whereas \textit{Output-based methods} score output tokens. \textbf{Bold} indicates the best result.}
\label{tab:main_tpr}
\renewcommand{\arraystretch}{1.2}

\resizebox{\textwidth}{!}{%
\setlength{\tabcolsep}{1.5mm}%
\begin{tabular}{@{}lccccccccccccc@{}}
\toprule

\multirow{2}{*}{\textbf{Method}}
& \multicolumn{4}{c}{\textbf{\textsc{Oracle-SFT}}}
& \multicolumn{4}{c}{\textbf{\textsc{Oracle-DPO}}}
& \multicolumn{4}{c}{\textbf{\textsc{Oracle-RL}}}
& \multirow{2}{*}{\textbf{Avg.}} \\

\cmidrule(lr){2-5}
\cmidrule(lr){6-9}
\cmidrule(lr){10-13}

& Math & Code & IF & QA
& Math & Code & IF & QA
& Math & Code & IF & QA
& \\

\midrule

\rowcolor{gray!8}
\multicolumn{14}{l}{
    \textcolor{gray}{\textbf{\textit{Input-based methods}}}
} \\

\addlinespace[2pt]

Perplexity
& 0.096 & 0.052 & 0.088 & 0.136
& 0.056 & 0.052 & 0.032 & 0.068
& 0.036 & 0.036 & 0.040 & 0.056
& 0.062 \\

Entropy
& 0.044 & 0.040 & 0.076 & 0.056
& 0.024 & 0.064 & 0.048 & 0.020
& 0.060 & 0.024 & \textbf{0.064} & \textbf{0.080}
& 0.050 \\

Lowercase
& 0.136 & 0.096 & 0.064 & 0.236
& 0.052 & 0.040 & 0.048 & 0.060
& 0.012 & 0.036 & 0.044 & 0.072
& 0.075 \\

Zlib
& 0.148 & 0.048 & 0.128 & 0.044
& 0.024 & \textbf{0.092} & \textbf{0.060} & 0.060
& 0.036 & 0.060 & 0.052 & 0.064
& 0.068 \\

Neighbor
& 0.057 & 0.071 & 0.061 & 0.085
& 0.063 & 0.019 & 0.031 & \textbf{0.107}
& 0.063 & 0.053 & 0.060 & 0.032
& 0.059 \\

\textsc{MIN-K\%}
& 0.100 & 0.060 & 0.116 & 0.172
& 0.056 & 0.052 & 0.016 & 0.064
& 0.044 & 0.040 & 0.036 & 0.056
& 0.068 \\

\textsc{MIN-K\%++}
& \textbf{0.200} & \textbf{0.144} & \textbf{0.232} & \textbf{0.404}
& \textbf{0.076} & 0.080 & 0.028 & 0.056
& \textbf{0.104} & 0.048 & 0.052 & 0.036
& \textbf{0.122} \\

\addlinespace[3pt]

\rowcolor{gray!8}
\multicolumn{14}{l}{
    \textcolor{gray}{\textbf{\textit{Output-based methods}}}
} \\

\addlinespace[2pt]

CDD
& 0.025 & 0.013 & 0.023 & 0.040
& 0.013 & 0.051 & 0.019 & 0.041
& 0.013 & 0.043 & 0.037 & 0.021
& 0.028 \\

Generated Perplexity
& 0.055 & 0.039 & 0.035 & 0.057
& 0.060 & 0.051 & 0.043 & 0.064
& 0.072 & 0.037 & 0.041 & 0.028
& 0.049 \\

Generated Entropy
& 0.049 & 0.060 & 0.037 & 0.065
& 0.064 & 0.041 & 0.033 & 0.068
& 0.069 & 0.037 & 0.036 & 0.036
& 0.050 \\

Generated \textsc{MIN-K\%}
& 0.053 & 0.048 & 0.033 & 0.064
& 0.067 & 0.051 & 0.041 & 0.061
& 0.071 & 0.040 & 0.044 & 0.028
& 0.050 \\

Min-kNN
& 0.049 & 0.000 & 0.000 & 0.029
& 0.057 & 0.000 & 0.000 & 0.000
& 0.065 & 0.035 & 0.000 & 0.000
& 0.020 \\

Self-Critique
& 0.107 & 0.075 & 0.016 & 0.059
& 0.032 & 0.089 & 0.035 & 0.033
& 0.041 & \textbf{0.091} & 0.016 & 0.047
& 0.053 \\

\bottomrule
\end{tabular}%
}
\end{table*}
\begin{table}[t]
    \centering
\caption{AUC of various detection methods across evaluation sets constructed with Math members and non-members from Math, Code, IF, or QA. Parentheses show relative AUC changes from the Math non-member baseline, with increases marked in red.}
    \label{tab:domain_shift}
    \renewcommand{\arraystretch}{1.2}
    \resizebox{0.9\textwidth}{!}{%
    \setlength{\tabcolsep}{8mm}%
    \begin{tabular}{lcccc}
        \toprule
        & \multicolumn{4}{c}{\textbf{Non-member Domain}} \\
        \cmidrule(lr){2-5}
        \textbf{Method} & \textbf{Math} & \textbf{Code}
                        & \textbf{IF} & \textbf{QA} \\
        \midrule
        Perplexity
        & 0.718
        & 0.920 \textcolor{red}{(+28.1\%)}
        & 0.955 \textcolor{red}{(+33.0\%)}
        & 0.968 \textcolor{red}{(+34.8\%)} \\

        Entropy
        & 0.944
        & 0.998 \textcolor{red}{(+5.7\%)}
        & 0.995 \textcolor{red}{(+5.4\%)}
        & 0.999 \textcolor{red}{(+5.8\%)} \\

        Lowercase
        & 0.775
        & 0.887 \textcolor{red}{(+14.4\%)}
        & 0.910 \textcolor{red}{(+17.4\%)}
        & 0.893 \textcolor{red}{(+15.2\%)} \\

        \textsc{MIN-K\%}
        & 0.707
        & 0.886 \textcolor{red}{(+25.3\%)}
        & 0.927 \textcolor{red}{(+31.2\%)}
        & 0.922 \textcolor{red}{(+30.5\%)} \\

        \textsc{MIN-K\%++}
        & 0.882
        & 0.921 \textcolor{red}{(+4.4\%)}
        & 0.924 \textcolor{red}{(+4.7\%)}
        & 0.893 \textcolor{red}{(+1.3\%)} \\

        \midrule
        \rowcolor{gray!15}
        Avg.
        & 0.805
        & 0.922 \textcolor{red}{(+14.5\%)}
        & 0.942 \textcolor{red}{(+17.0\%)}
        & 0.935 \textcolor{red}{(+16.1\%)} \\
        \bottomrule
    \end{tabular}}
\end{table}

\end{document}